\documentclass[10pt,conference]{ieeeconf}

\usepackage{newclude}
\usepackage{graphicx}
\usepackage{color}
\usepackage{subfigure}
\usepackage{float}
\usepackage{paralist}
\usepackage{mathtools}
\usepackage{siunitx}
\usepackage{caption}

\usepackage{algorithm}
\usepackage{algpseudocode}

\usepackage{amsthm}
\usepackage{amsmath}
\usepackage{amssymb}
\usepackage{amsfonts}

\IEEEoverridecommandlockouts
\definecolor{mgnote}{rgb}{1,0,0}

\definecolor{pnnote}{rgb}{0,0,1}

\title{
    \LARGE \bf
    The Data Market: Policies for Decentralised Visual Localisation
}

\author
{
    Matthew Gadd and Paul Newman
    \thanks
    {
        Authors are from the Oxford Robotics Institute, University of Oxford, Oxford, England. 
        {
            \tt\small \{mattgadd,pnewman\}@robots.ox.ac.uk
        }
    }
}

\begin{document}
    \maketitle
    \thispagestyle{empty}
    
    \begin{abstract}
\label{secs:abstract}
This paper presents a mercantile framework for the decentralised sharing of navigation expertise amongst a fleet of robots which perform regular missions into a common but variable environment.
We build on our earlier work \cite{GaddIROS2016} and allow individual agents to intermittently initiate trades based on a real-time assessment of the nature of their missions or demand for localisation capability, and to choose trading partners with discrimination based on an internally evolving set of beliefs in the expected value of trading with each other member of the team.
To this end, we suggest some obligatory properties that a formalisation of the distributed versioning of experience maps should exhibit, to ensure the eventual convergence in the state of each agent's map under a sequence of pairwise exchanges, as well as the uninterrupted integrity of the representation under versioning operations.
To mitigate limitations in hardware and network resources, the ``data market'' is catalogued by distinct sections of the world, which the agents treat as ``products'' for appraisal and purchase.
To this end, we demonstrate and evaluate our system using the publicly available \textit{Oxford RobotCar Dataset} \cite{RobotCarDatasetIJRR}, the hand-labelled data market catalogue (approaching \SI{446}{\km} of fully indexed sections-of-interest) for which we plan to release alongside the existing raw stereo imagery.
We show that, by refining market policies over time, agents achieve improved localisation in a directed and accelerated manner.
\end{abstract}
    \section{Introduction}
\label{secs:introduction}

Concerted action amongst a fleet of robots is a tangibly effective strategy for executing a common task or set of tasks (e.g. mapping, exploration, target tracking, path planning, surveillance), particularly under realistic considerations of resource limitations (e.g. bandwidth, computational, storage).
Mankind has refined its approach to similar problems over millenia, resulting in today's sophisticated market economies in which global production is optimised by the redistribution of resources and responsibilities to individuals, who are themselves profit-seeking agents.

To date, roboticists have effectively dispatched lifelong navigation systems \cite{milford2012seqslam,naseer2014robust} to deal with stark scene change.
The consequential explosion in experience density required to capture complex spectra of change (and the associated processing load) are mitigated by strategies including: summarised, reduced map representations \cite{dymczyk2015gist}, prioritising the utility of past experiences to predict the real-time relevance of memories \cite{linegar2015work}, and induced forgetfulness \cite{milford2010persistent}.

In this paper, we contribute to this conversation by:
\begin{itemize}
	\item Posing and modelling the sharing in the fleet as an instance of distributed version control, abstracting data fusion as a commutative operation which guarantees synchronicity in representation amongst the team, and the avoidance of corruption, Sections \ref{secs:preliminaries} and \ref{secs:branching},
	\item Providing a candidate set of policies which agents can use as a ``pricing'' strategy to explore the marketplace, based on an evolving set of beliefs in the likely boon after making a purchase from another identifiable member of the team, Section \ref{secs:market}, and
	\item Transforming the team's operational envelope into this marketplace on the basis of demarcating the data that can be collected within it over the lifetime of operation of the fleet, Section \ref{secs:products}.
	\item Suggesting an abstracted, middleware-specific, shared finite-state machine which can be used as both the medium for market activity and for exactly repeatable simulations of arbitrarily large fleets, which is useful for investigating the effect of a heterogeneous landscape of algorithms present in the market, Section \ref{secs:state_machines}.
\end{itemize}

While our framework is agnostic to the nature of (or quantification of the value of) data to be shared, we leverage it within the context of lifelong visual localisation of a fleet of vehicles in a city-scale urban setting, serving as both an application poignant to the community, and a pedagogical example through which to develop the versioning and pricing strategies, and show in Section \ref{secs:experiments} that the targeted sharing of navigation expertise provides ever-increasing robustness against workspace dynamics.
    \section{Related Work}
\label{secs:related_work}

Multi-robot, keyframe-based SLAM is serviceable by estimation of the relative position of robots in a central frame of reference grown by the fusion of individual maps \cite{riazuelo2014c}.
The problem is further divorced from the estimation algorithm by the design of data-structures and processes for thread-safe modification of the central map by local maps constructed with generalised egomotion estimation and place recognition, but still relies on a central map assembly platform \cite{forster2013collaborative}.

Distributed estimation strategies are a means of escaping the central authority, \cite{carrillo2013decentralized}, but are subject to a unimodal distribution of the state estimate as a result of the required linearisation, which is addressed by probabilistic modelling of pairwise interactions and used by the team to motivate exchanges \cite{fox2000probabilistic}. 
Ceilings in communication range and bandwidth are surpassed by marginalising agent-specific factors from batch-summarised local-maps before sharing \cite{cunningham2013ddf}.
Data summarisation, however, preempts individually held but globally consistent maps, addressed by ``torrent'' tools for web-based distributed map management \cite{cieslewski2015map}.

Indeed, approaches to collaborative localisation (CL) are increasingly borrowing from the vocabulary of software versioning and management \cite{cieslewski2017efficient}.
In centralised version control systems (VCS) a master repository maintains a history of all incremental changes and consolidates the changes from subordinates \cite{nagel2005subversion}.
Decentralised systems, however, dictate that suboordinates own their own copies of the repository, where updates can take place in the form of any number of interactions between any combination of agents \cite{loeliger2012version}.

A combinatorial space of interactions of this kind can be explored by role-assignment within a team behaving under common auction policies \cite{frias2004exploring}.
Auctions, however, require some measure of the quality of trades, which is facilitated in \cite{wang2016pricing} by the use of a pricing model in a leader-follower regime.
    \section{Preliminaries}
\label{secs:preliminaries}

Robots collaborating for the shared objective of mapping and localising within a variable world will record and share an unordered set of ``patch'' updates, which need to be communicated and incorporated in a consistent manner.
In Section \ref{secs:branching}, we will put forward a formalisation of mapping through experience in terms of version control operations.
To this end, we begin by drawing on the notation of \cite{dagit2009type} to explain some basic elements of the \textit{Patch Theory} used in evaluating correctness for the Darcs VCS \cite{roundy2005darcs}.

\begin{figure}
    \centering
    \includegraphics[width=0.25\textwidth]{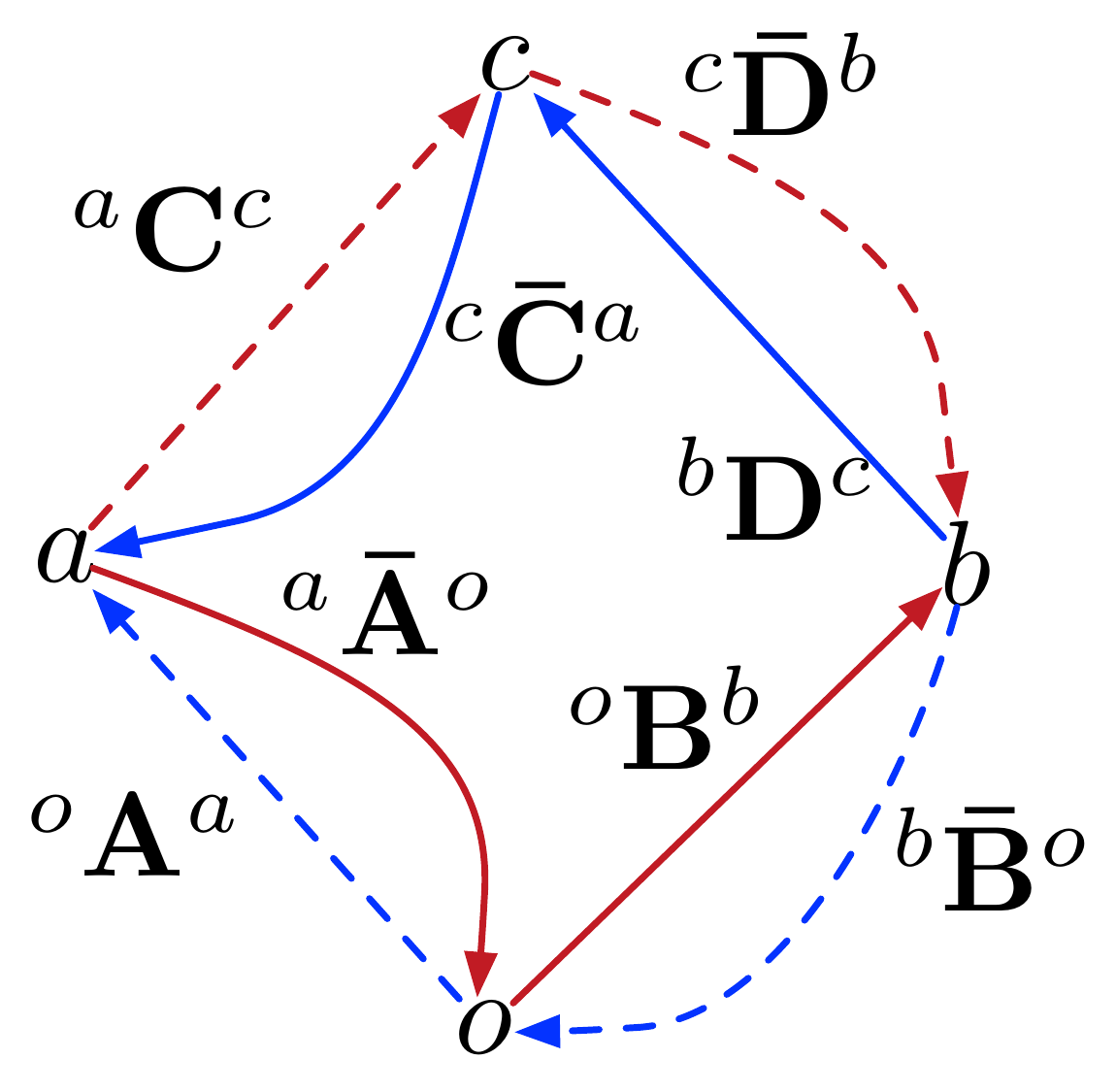}
    \caption{In general, a collection of unordered divergent patches can be consistently consolidated if the input and output states are monitored carefully, and by the design of a ``commutation policy'', $\Gamma$, which must take into account the underlying nature of the data which is being versioned.}
    \label{figs:preliminaries:commuting}
\end{figure}

\subsection{Repository states}
\label{secs:preliminaries:states}

A repository, $\mathcal{R}$, tracks the incremental changes to a working copy of some data, the current state of which can be uniquely identified, $\mathcal{R}^{a}$. 

\subsection{Patch updates}
\label{secs:preliminaries:patch_updates}

A patch, $^{a}\mathbf{A}^{b}$, is a concrete representation of a transformation made to the state of the repository, identified by input and output states, $a$ and $b$.

\subsection{Patch equality}
\label{secs:preliminaries:patch_equality}

Given two patches, $^{a}\mathbf{A}^{b}$ and $^{c}\mathbf{B}^{d}$, we define the patches as describing the same transformation of state if  $d$ is equivalent to $b$ whenever $b$ is equivalent to $a$.

\subsection{Patch composition}
\label{secs:preliminaries:patch_composition}

Patches with coincidental output and input states, $^{a}\mathbf{A}^{b}$ and $^{b}\mathbf{B}^{c}$, can be combined in sequence, and written as $^{a}\mathbf{A}^{b}\mathbf{B}^{c}~=~^{a}\mathbf{AB}^{c}~=~^{a}\mathbf{C}^{c}$.

\subsection{Patch inverse}
\label{secs:preliminaries:patch_inverse}

If $^{a}\mathbf{A}^{b}$ is a patch, and $^{c}\mathbf{\bar{A}}^{d}$ is another patch with the opposite transformation of state, we require that for every equivalence made between states $d$ and $a$, states $c$ and $b$ are also equivalent.

\subsection{Patch commutivity}
\label{secs:preliminaries:commutation}

Two patches in sequence, $^{o}\mathbf{A}^{a}$ and $^{a}\mathbf{B}^{b}$, are said to commute to some other pair, $^{o}\mathbf{B}_{1}^{a}$ and $^{a}\mathbf{A}_{1}^{b}$, if they have the same effect in any sequence, $^{o}\mathbf{A}\mathbf{B}^{b}~\leftrightsquigarrow~^{o}\mathbf{B}_{1}\mathbf{A}_{1}^{b}$ and $^{o}\mathbf{B}_{1}\mathbf{A}_{1}^{b}~\leftrightsquigarrow~^{o}\mathbf{A}\mathbf{B}^{b}$ (where $\leftrightsquigarrow$ indicates that the sequences are related by some number of commutes).

\subsection{Merging divergent patches}
\label{secs:preliminaries:merging}

Consider in Figure \ref{figs:preliminaries:commuting} a pair of divergent (equivalent input states) patches $^{o}\mathbf{A}^{a}$ and $^{o}\mathbf{B}^{b}$, and some candidate patches, $^{a}\mathbf{C}^{c}$ and $^{b}\mathbf{D}^{c}$, the action of which should be convergent, $^{o}\mathbf{A}\mathbf{C}^{c}~\leftrightsquigarrow~^{o}\mathbf{B}\mathbf{D}^{c}$.

These commutative pairs are illustrated in Figure \ref{figs:preliminaries:commuting}, and the guarantee of a successful merge given their existence can be understood visually by following the effects of patches and their inverses between common states $o$ and $c$.

Here, solid paths indicate ``patches being commuted'', and dashed paths indicate ``patches commuted to''.

In short, if there is some symmetric ``commutation policy'', $\Gamma$, such that $(^{a}\mathbf{C}^{c},~^{c}\mathbf{\bar{D}}^{b}) = {\Gamma}(^{a}\mathbf{\bar{A}}^{o},~^{o}\mathbf{B}^{b})$ (red paths) and $(^{b}\mathbf{D}^{c},~^{c}\mathbf{\bar{C}}^{a}) = {\Gamma}(^{b}\mathbf{\bar{B}}^{o},~^{o}\mathbf{A}^{a})$ (blue paths), and the reverse effect of patch effects can be characterised with high fidelity, then we can always merge divergent patches.
    \section{A Distributed Branching Model for Experience Maps}
\label{secs:branching}

Any application of the theory presented in Section \ref{secs:preliminaries} to multirobot map management will, at the very least, need to
\begin{enumerate}
    \item Delineate map updates as incremental patches which can be applied (whether the nature of the application is constructive, or destructive) to the repository, as discussed in Section \ref{secs:preliminaries:patch_updates}
    \item Instantiate a valid commutation policy which can consolidate divergent representations, as discussed in Sections \ref{secs:preliminaries:commutation} and \ref{secs:preliminaries:merging}
\end{enumerate}

\subsection{Experience-based navigation}
\label{secs:branching:experience_maps}

\begin{figure}
    \centering
    \includegraphics[width=0.3\textwidth]{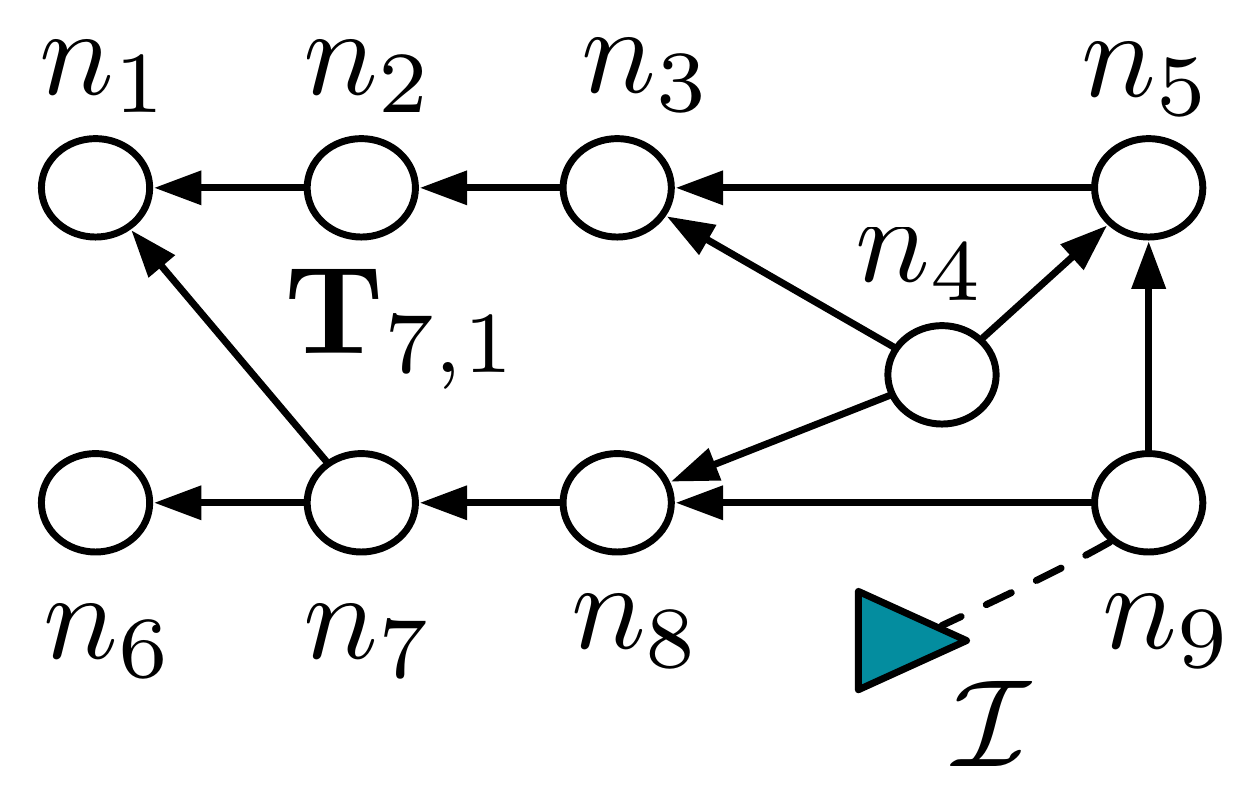}
    \caption{Experience maps are implemented as a database of nodes, $n$, connected by directed edges, which define topometric neighbourhoods formed from 6DoF relative poses, $\mathbf{T}$. This work relies on the EBN package developed in \cite{LinegarICRA2015}, but is applicable to management of experience maps in general.}
    \label{figs:branching:experience_maps}
\end{figure}

Due to complex modes of change in the environment's appearance (e.g. weather, illumination, structure), visual localisation is difficult.
Our robots have typically approached this problem by gathering overlapping representations of the world, when the state of the map is not sufficient to explain the robot's current experience of a place (i.e. when localisation fails) \cite{LinegarICRA2015,GaddICRA2015}.

As shown in Figure \ref{figs:branching:experience_maps}, experience maps use topometric graphs $\mathcal{G}$, or relational databases, where nodes may store a characterisation of places (e.g. 3D landmarks) and edges store rigid relationships between these places (e.g. six degree of freedom (6DoF) relative poses).

Localisation is achieved by searching (e.g. breadth-first) local graph neighbourhoods, with a seed that is provided by a search in apperance space (e.g. FAB-MAP \cite{Cummins2010}), and is considered successful on a sufficiently robust match (e.g. by inlier count) to some previously recorded map content.

\subsection{Experience repositories}
\label{secs:branching:repository}

Each agent, $r_{i}$, in the fleet possesses a repository $\mathcal{R}_{i}^{a} = \{\mathcal{G}_{i}^{a},~^{o}\mathcal{H}_{i}^{a}\}$ where the map $\mathcal{G}_{i}^{a}$ is in a state $a = \{p~\vert~p = (\iota, g), \iota = 1,~g = (n, \mathcal{E}),~n \in \mathcal{G}_{i}^{a},~\mathcal{E} = \{e~\vert~e \in \mathcal{G}_{i}^{a}\}\}$ summarising its content and connectivity (uniquely identified across the team's databases using $128$-bit UUIDs).

The history of the repository is encoded in a sequence of patches $^{o}\mathcal{H}_{i}^{a} = \{^{o}\mathbf{P}_{i}^{a}, \ldots\}$.
Considering our singular goal of improving localisation, we treat the sequence, $^{o}\mathcal{H}_{j}^{a}$, as a linear history, with no support for multiple feature branches local to each robot, and as such our system is more similar to Darcs \cite{roundy2005darcs} than Git \cite{loeliger2012version}.

\subsection{Recording changes to the working copy}
\label{secs:branching:recording}

When a robot, $r_{i}$, executes a foray into the world, the novel content it records is represented as a patch $^{a}\mathbf{P}_{i}^{b}$ consisting of a set of atomic patch elements $p = \{\iota , g\} \in~^{a}\mathbf{P}_{i}^{b}$, each of which either inserts ($\iota = 1$) or deletes ($\iota = 0$) a node and its out-edges, $g = (n, \mathcal{E})$, such that the map changes from state $a$ to state $b$ as $\mathcal{G}_{i}^{b} = \mathcal{G}_{i}^{a}\mathbf{P}_{i}^{b},~{\vert}\mathcal{G}_{i}^{a}{\vert} \geq 0$.
For ease of notation, we will refer to states and patches using only nodes in the sets, i.e. $n \in ^{a}\mathbf{P}_{i}^{b}$

\subsection{Merging traded mapping data}
\label{secs:branching:merging}

\begin{figure}
    \centering
    \includegraphics[scale=0.3]{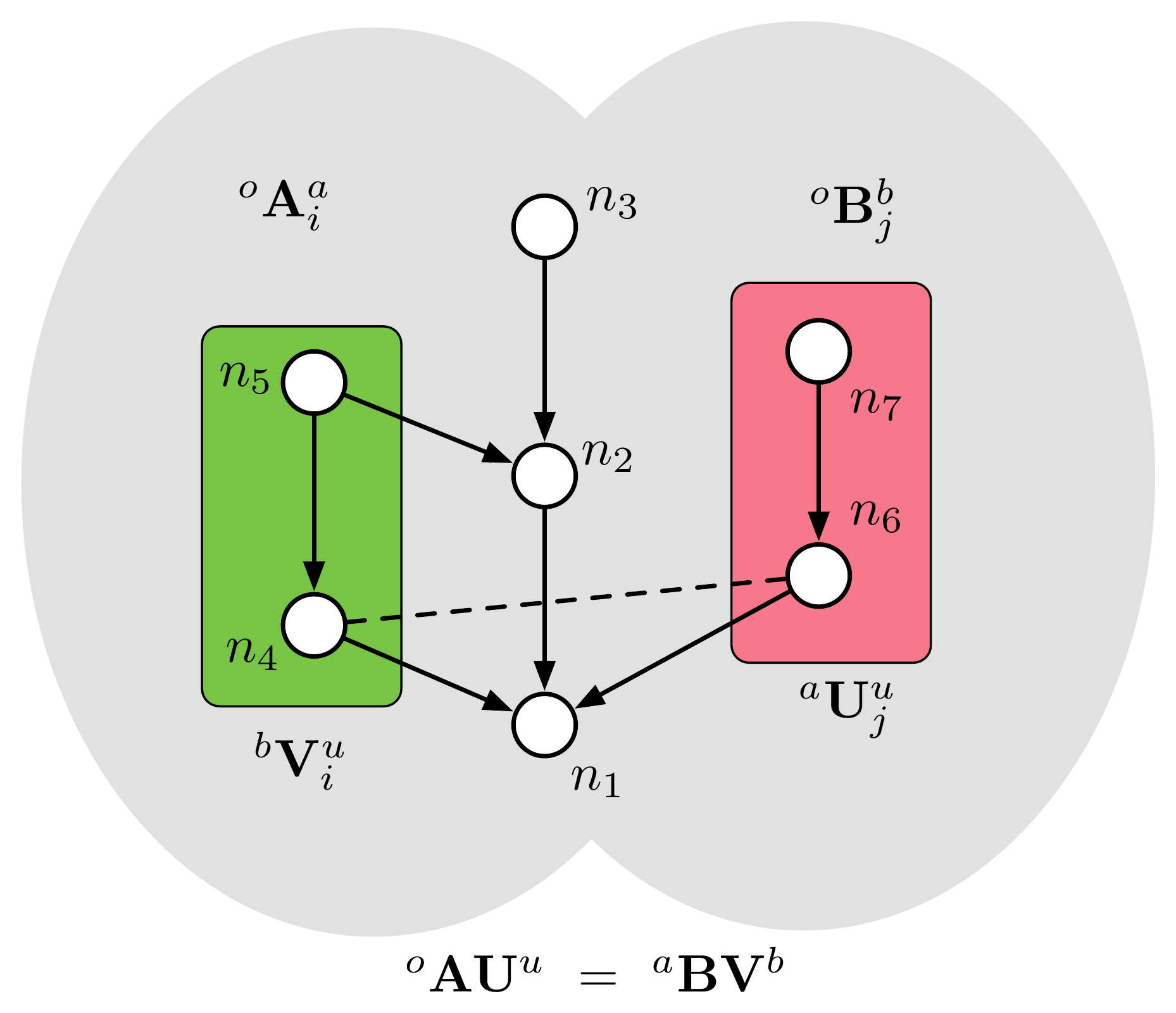}
    \caption{A pairwise exchange of patches by agents with some common repository history, but currently divergent map representations. To guarantee symmetry and consistency in such an exchange, and particularly over the lifetime of a fleet of arbitrary size, $R$, the robots must share a common understanding of how to commute divergent content, $\Gamma_{i} = \Gamma~\forall~i \in R$, as well as agree on a common protocol for valuing, choosing between, and discarding from closely matching content, $\gamma_{i} = \gamma~\forall~i \in R$.}
    \label{figs:branching:sets}
\end{figure}

In Figure \ref{figs:branching:sets}, we illustrate the states of two robots' repositories, $\mathcal{R}_{i}^{a}$ and $\mathcal{R}_{j}^{b}$, at the point of initiating a trade.
The robots share an original common state $o = \{n_{1}, n_{2}, n_{3}\}$ (they have traded with each other before), but (through either trading with other agents, or by virtue of having mapped the world recently) have diverged in their representation, $^{o}\mathbf{A}_{i}^{a} = \{n_{1}, n_{2}, n_{3}, n_{4}, n_{5}\}$ and $^{o}\mathbf{B}_{j}^{b} = \{n_{1}, n_{2}, n_{3}, n_{6}, n_{7}\}$.
Consider two candidate convergent patches $^{b}\mathbf{V}_{i}^{u} = \{n_{4}, n_{5}\}$ and $^{a}\mathbf{U}_{j}^{u} = \{n_{6}, n_{7}\}$, which would transform each repository to a common state $u = \{n_{1}, \ldots, n_{7}\}$, where $^{a}\mathbf{U}_{j}^{u} = ^{o}\mathbf{B}_{j}^{b}{\setminus}^{o}\mathbf{A}_{i}^{a}$ and $^{b}\mathbf{V}_{i}^{u} = ^{o}\mathbf{A}_{i}^{a}{\setminus}^{o}\mathbf{B}_{j}^{b}$ are equivalent to a \textit{diff} operation in text-based version control.

A perfectly reasonable and simple strategy for commuting divergent patches is the set union, $\Gamma_{\cup}$, or in this example $^{a}\mathbf{C}_{j}^{c} = ^{b}\mathbf{V}_{i}^{u}$ and $^{b}\mathbf{D}_{i}^{c} = ^{a}\mathbf{U}_{j}^{u}$.

However, in the spirit of experience-based navigation, robots should not add content to their maps which can be sufficiently explained by existing map content, as this would result in ever-increasing experience density, along with the associated bottlenecks in performance.
To this end, we advocate an alternative commutation policy, $\Gamma_{\mathcal{M}}$, which uses the underlying localiser itself to search for pairwise matches $\mathcal{M}_{i, j} = \{(n_{i}, n_{j})~\vert~n_{i} \in ^{o}\mathbf{V}_{i}^{a},~n_{j} \in ^{o}\mathbf{U}_{j}^{b}\}$ between divergent patches.

Disregarding closely matching content in this way must be expressed through versioning operations as either
\begin{inparaenum}[(i)]
	\item Keeping content already in the database, and ignoring the matched content offered through the trade, or
	\item Choosing the exchanged content, and removing the original content from the working copy, and repository history.
\end{inparaenum}
Each of these considerations can be problematic in their own way, and we discuss the related issues in Sections \ref{secs:branching:symmetry} and \ref{secs:branching:integrity}.

\subsection{Maintaining symmetry in versioning operations}
\label{secs:branching:symmetry}

In order to maintain symmetry in the maps of various agents in the team, $\Gamma_{\mathcal{M}}$ relies on a repeatable and universal (to the team) ``choice policy'' for deciding which of two pairwise matching elements should be preferred, $n^{+}~\leftarrow~{\gamma}(n_{i}, n_{j})$, and which should be disregarded, $n^{-}~\leftarrow~{\gamma}^{-1}(n_{i}, n_{j})$.

Some choice policies which would clearly cause mischief in versioning might include:
\begin{inparaenum}[(i)]
    \item consistent selection of the left-hand-side (LHS) content, or
    \item random selection based on a the toss of a coin,
\end{inparaenum}
However, as we will discuss further in Section \ref{secs:market:price}, $\gamma$ is inherently related to the quality of the map, and as such to the performance of the localiser.
As such, in this work, we have investigated selection mechanisms which include: 
\begin{inparaenum}[(i)]
    \item the quality of egomotion estimation (stereo match inlier counts) on the map frames,
    \item appearance similarity (FAB-MAP match scores from candidate to map), and
    \item the number of times the localiser has matched to the candidate node, using the concept of ``path memory'' presented in \cite{LinegarICRA2015}.
\end{inparaenum}

\subsection{Maintaining map integrity under patch updates}
\label{secs:branching:integrity}

\begin{figure*}
    \centering
    \includegraphics[scale=0.25]{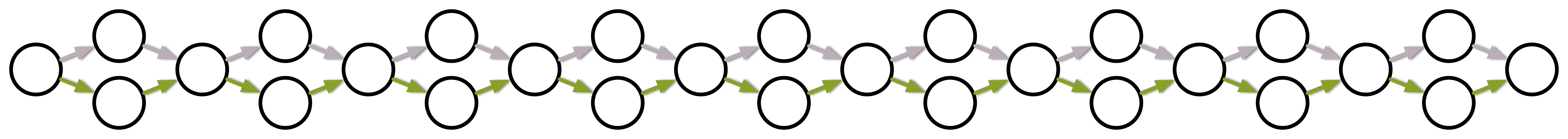}
    \caption{Mutual representation of repository history for a team of robots ($R = 2$) which are repeatedly mapping the world, and sharing all mapping information at every opportunity ($K = 9$). Over the course of all trials ($M = 100$), there is no deviation from proper convergence.}
    \label{figs:experiments:convergence:dm_dot}
\end{figure*}

The localiser-inspired commutation we advocate, $\Gamma_{\mathcal{M}}$, must do additional work to maintain the integrity of the experience map, $\mathcal{G}$, to avoid errata in the form of disconnected subgraphs (which can be catastrophic for motion planners).

In general, given any arbitrary collection of divergent patch pairs $\mathcal{C} = \{\mathcal{C}_{i} = (^{o}\mathbf{A}_{i}^{a},~^{o}\mathbf{B}_{j}^{b})\}$, we could hypothetically design a battery of tests $\mathcal{T} = \{\mathcal{T}_{j}\}$, each of which indicates the correctness of $\Gamma$ as it acts on each node $n \in \mathcal{C}_{i}$, $\mathcal{T}_{j}(n) \in \{0, 1\}$.

The result of each of the tests on a particular node gives us a binary test-vector $\mathcal{T}(n) = [\mathcal{T}_{1}(n), \mathcal{T}_{2}(n), \ldots, \mathcal{T}_{{\vert}\mathcal{T}{\vert}}(n)]^{T}$, and the results of the entire test-battery constitute multiset of test-vectors over the configuration $\mathcal{T}(\mathcal{C}_{i}) = [\mathcal{T}(n_{1}), \ldots, \mathcal{T}(n_{{\vert}\mathcal{C}{\vert}})]$.

Now, all tests over all nodes in every candidate configuration gives us the multiset union $\mathcal{T}(\mathcal{C}) = \bigcup\mathcal{T}(\mathcal{C}_{i})$.

If the cardinality of this multiset, $\operatorname{card}(\mathcal{T}(\mathcal{C})) = 2^{J}$, the collection $\mathcal{C}$ is considered to have sufficiently tested the battery $\mathcal{T}$.

More specifically, considering the effect of the choice policy on the database, $n^{+}~\leftarrow~{\gamma}(n_{i}, n_{j})$ and $n^{-}~\leftarrow~{\gamma}^{-1}(n_{i}, n_{j})$, some useful tests in the battery may include:
\begin{inparaenum}[(i)]
	\item checking that $n^{+}$ maintains connectivity with all of the adjacent nodes of $n^{-}$, or
	\item $n^{+}$ and $n^{-}$ never appear alongside one another in the database schema,
\end{inparaenum}
while the configurations $C_{i}$ may be randomly generated, or may be parsed from exemplar mapping data via the navigation framework.

\subsection{Randomised commutation verfication}
\label{secs:verification}

\begin{figure}
    \centering
    \includegraphics[scale=0.25]{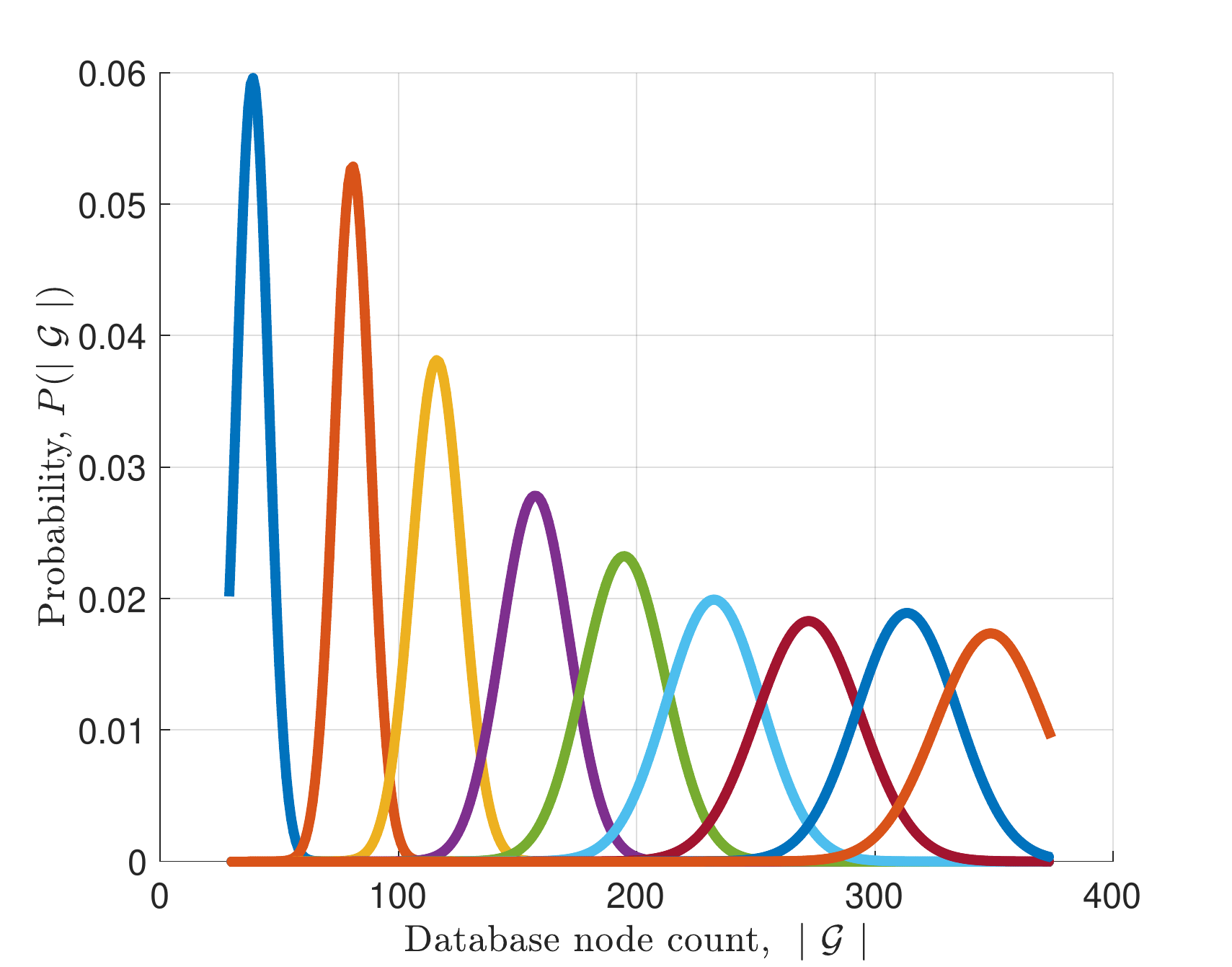}
    \caption{Distribution of node counts at each convergence basin $0 < k < 9$, with lengthening tails at later stages in the simulation.}
    \label{figs:experiments:convergence:num_rndf_node_ids}
    \vspace{-.6cm}
\end{figure}

It is beyond the scope of this paper to suggest an irrefutable test battery or enumerate every corner-case collection of divergent patches, but we show in Figure \ref{figs:experiments:convergence:dm_dot} the mutual histories (see Section \ref{secs:branching:repository}) aggregated over $M$ rounds of \textit{Monte Carlo} \cite{robert2004monte} analysis within a team of $R$ robots as they engage in $K$ forays, where each foray instigates toy mapping data from every robot, $r_{i}$, the magnitude of which is normally distributed ${\vert}\mathbf{P}_{i, k}{\vert} \sim \mathcal{N}(\mu_{i}, \sigma_{i}^{2})$, and results in a wholesale trade between all pairs of agents (i.e. a \textit{distributed-query-all} trading strategy).
We also aggregate, in Figure \ref{figs:experiments:convergence:num_rndf_node_ids}, the node count at every point of convergence, where it can be seen that the expected database size at convergence point $k$ is $E[{\vert}\mathcal{G}_{i, k}{\vert}] = k\sum\mu_{i}$, whereas the variance will also have grown as $E^{2}[{\vert}\mathcal{G}_{i, k}{\vert}] = k\sum\sigma_{i}^{2}$.
Aside from justifying our confidence in the integrity of our commutation policy, ${\Gamma}_{\mathcal{M}}$, this conveniently illustrates two points:
\begin{itemize}
    \item Trading with other robots introduces uncertainty into the map representation, and
    \item Without effective strategies for \textit{throttling} data exchanges, maps will grow unreasonably
\end{itemize}
We show how our ``data market'' can mitigate these concerns in Sections \ref{secs:market} and \ref{secs:products} respectively.

    \section{A Data Market}
\label{secs:market}

Section \ref{secs:branching:symmetry} introduced the choice policy $\gamma$, the use of which is not limited to versioning, and can be expanded upon to characterise a ``price'' for patches, $\mathbf{P}_{i, k}$, as recorded individually by an agent $i$ and entered into a ``data market'' (DM) for trading at time $k$.

\subsection{Pricing data transactions}
\label{secs:market:price}

Specifically, because $\gamma$ distinguishes between nodes as a selection mechanism, it can be repurposed to act as a metric for their value, ${\gamma}(n) \geq 0~\forall n \in \mathbf{P}_{i, k}$.
The implication of this value to a robot in the DM may be:
\begin{itemize}
    \item The \textit{cost} of the transaction (e.g. packet size, on limited bandwidth networks)
    \item Some \textit{reward} associated with the transaction (e.g. any of the candidate policies suggested in \ref{secs:branching:symmetry}, and their relationship to localisation performance)
\end{itemize}

Thus, on applying a patch received from the trading partner, $r_{j}$, each robot, $r_{i}$, makes a measurement of the value of the transaction as ${m}_{j, k} = \frac{1}{\vert~\mathbf{P}_{j, k}~\vert}\sum_{n \in \mathbf{P}_{j, k}}{\gamma}(n)$, normalised by the size of the patch, and thus expressed as a per-packet-price.
The robot keeps a record of these measurements, $M_{j} = \{m_{j, 0}, \ldots, m_{j, k}\}$.

\subsection{Tracking trust in marketplace vendors}
\label{secs:market:beliefs}

Each robot, $r_{i}$, manages an internal belief in the expected value of performing a transaction with another robot, given the opportunity $E[m_{j, k}] = \frac{1}{{\vert}M_{j}{\vert}}\sum_{i = 0}^{k - 1}m_{j, i}$, modelled as a normally distributed random variable $M_{j} \sim \mathcal{N}(\mu_{j}, \sigma_{j})$.

To avoid repeated (perhaps expensive) computation of trade values, we track the mean and variance as ${\mu}_{j, k} = {\mu}_{j, k - 1} + \frac{1}{{\vert}M_{j}{\vert}}({m}_{j, k} - {\mu}_{j, k - 1})$ and ${\sigma}_{j, k} = {\sigma}_{j, k - 1} + (m_{j, k} - {\mu}_{j, k - 1})(m_{j, k} - {\mu}_{j, k})$ respectively, using the latest measurement as an observation, $m_{j, k}$.

\subsection{Choosing trading partners with discrimination}
\label{secs:market:choosing}

At any point in time, a robot interested in trading sends a down-sampled patch representation of its newly recorded content to every candidate seller, $\mathbf{S}_{i, k} = \mathcal{S}(\mathcal{G}_{i}),~{\vert}N{\vert} \ll {\vert}\mathcal{G}_{i}{\vert}$, where the ``sampling policy'', $\mathcal{S}$, is a nod to bandwidth limitations, and must rank candidate packets by their price as evaluated by $\gamma$, and adhere strictly to limits imposed by the medium over which samples are sent.

Team members who are available to trade take the liberty of performing measurements on this sample, ${m}'_{j}$, described in Section \ref{secs:market:price}, which the buyer collects as ``tender'' offers, $\boldsymbol{m'} = \{{m}'_{j}~\vert~j \in R,~j \neq i\}$, in a tender-adjudication style process.

The buyer then chooses the seller for which the measurement is least perturbed from its internal beliefs, $j^{*} = \operatorname{min}_{j \in R}~\vert~{m}'_{j} - E[m_{j}]~\vert$, before executing an up-sampled trade with $j^{*}$ and updating its trust in that seller, as described in Section \ref{secs:market:beliefs}.

\subsection{Exploring the marketplace by potential value}
\label{sections:market:bandit}

The tender-adjudication process described in Section \ref{secs:market:choosing} can be thought of as an observation of the reward after pulling an arm corresponding to each team member in a multi-armed bandit \cite{auer2002nonstochastic}
Over the course of $K$ such trades, we are trying to maximise the expected total reward $ \sum_{k \in K}~E[\vert~{m}'_{j, k} - E[m_{j}]~\vert]$.
We apply an epsilon-greedy approach \cite{langford2008epoch} to the selection of trading partner, either exploiting the best lever (the seller with the best expected normalised patch price) or exploring (choosing a random trading partner) for fractions $\epsilon$ and $1 - \epsilon$ of the total trials respectively.
However, the world is vast, and its appearance is subject to complex modes of change, and it may be that a particular agent consistently provides higher quality content from one part of the environment (e.g. it has mapped that environment more regularly).
    \section{A Marketplace Catalogue Protocol}
\label{secs:products}

Indeed we continue maturing our ``data market'' by cataloguing the world as a ``product range'' which the fleet can use to understand both the world, and the relative value that their peers offer to them.
We propose the use of a ``product policy'' which can approximately demarcate database content into sections-of-interest, $\mathcal{P} = \{p_{i}~\vert~i \in {\vert}\mathcal{P}{\vert}\}$, which may correspond to unique streets or local attractions along the team's typical operating envelope.
Under this framework, robots now act as vendors for a range of commodities in the marketplace.

\subsection{Weak localisers as product policies}
\label{secs:products:policy}

Some candidates for this policy may include:
\begin{inparaenum}[(i)]
	\item Global positioning system (GPS), or
	\item clusters of visual words which are common to certain scenes, or
	\item laser scan matching for local geometry.
\end{inparaenum}
For the purposes of an approximate ground-truth, we have augmented our \textit{Oxford RobotCar Dataset} \cite{RobotCarDatasetIJRR} with a \textit{MATLAB} graphical toolkit for labelling vast collections of video imagery with a direct lookup from frame-index to street name\footnote{In this work, we have created and used labels for approximately \SI{446}{\km} of such data, which we plan to publish at \url{http://robotcar-dataset.robots.ox.ac.uk} in a convenient format (e.g. XML, YAML, JSON) along with a set of scripts for manipulating the vast scale of sensor records in the \textit{Oxford RobotCar Dataset} \cite{RobotCarDatasetIJRR}.}.
Thus, each member of the team holds a common topology representing sections of the environment and their connectedness.
This topology is distinct from the database schema which each robot is populating by mapping exercises, and in which localisation is performed (see Section \ref{secs:branching:experience_maps}).

\subsection{Itemised wares}
\label{secs:products:items}

Each agent can use $\mathcal{P}$ to curate a record of which nodes belong to distinct sections $\mathcal{L} = \{L_{i}~\vert~L_{i} = \{n \in p_{i} \}\}$ such that its current repository state is $\mathcal{G} = \cup_{p_{i}}L_{i}$.

Additionally, we extend the tracked beliefs introduced in Section \ref{secs:market:beliefs} and by including a record of the nodes (identifiers only) used in expectation updates such that $\mathbf{M}_{j} = \{\mathbf{m}_{j, k}~\vert~\mathbf{m}_{j, k} = (\mathbf{P}_{j, k}, m_{j, k})~\vert\}$, which can be related to a purchase record for foreign content received from team members, $\mathcal{A} = \{A_{i}~\vert~A_{i} = \{n \in p_{i} \cap \mathbf{m}_{j, k} \}\}$.

Finally, it will prove useful to track the popularity of products as sold by an agent acting as a vendor in the marketplace, $\mathcal{S} = \{S_{i}~\vert~S_{i} = \{n \in p_{i},~n \notin \mathcal{A}\}\}$, as an apparent measure of trust that other team members have in the value of the packets it has entered into the market.

\subsection{Flexible and targeted shopping lists}
\label{secs:products:shopping}

Equipped with this more nuanced understanding of its environment, an agent embarking on a foray into the world, can use $\mathcal{P}$ as a weak indication of its position within the catalogue, $p^{'} = \mathcal{P}(\mathcal{I})$, where the current frame, $\mathcal{I}$, must be serialised to contain any data that $\mathcal{P}$ might require in its lookup (e.g. bag-of-words, noisy INS traces, etc).

Then, the most simpleminded shopper might purchase the current product: $p_{k} = p^{'}$.

However, robots are equipped with a variety of noisy sensors, and a more robust shopping strategy would be querying for products which are close to each other in product-space, $p_{k} = \{\ldots, p^{'-1}, p^{'}, p^{'+1}, \ldots\}$, which should account for variation in metric understanding of the world across the team.

However, agents operating in our ``data market'' are allowed to share advisories with each other, which indicate their favourite sellers for a particular product $j^{*} = \operatorname{max}_{j \in R}(E[m_{j, k}])$, as well as advertisements, which indicate their own best-selling products $p^{*} = \operatorname{max}_{i \in {\vert}\mathcal{S}{\vert}}({\vert}S_{i}{\vert})$, which can be leveraged in a buyer-recommendation style decision-making process to occasionally shop for the best-selling item from an overall favourite seller for the current product.
    \section{A Multirobot Finite-State Machine}
\label{secs:state_machines}

In the interests of performing exactly-repeatable simulations over many trials, we have developed a middleware extension to our ``Mission-Oriented Operating Suite'' (MOOS) \cite{newman2008moos}, with a public interface that guarantees synchronisation for teams of robots which are alternatively recording and sharing mapping data as discussed in Section \ref{secs:market} and \ref{secs:products}.

We consider each robot as a finite-state machine (FSM) \cite{bosik1991finite}, which must publish only a state $\boldsymbol{\theta}_{i, k} = (\theta_{i, k}, \psi_{i, k})$ where $\theta_{i, k}$ is a state descriptor (e.g. ``mapping'', ``sampling'', ``purchasing'') and $\psi_{i, k}$ is a semaphore primitive \cite{kosaraju1973limitations} which counts iterations of each state-type and provides entry and exit synchronisation points for agents which may have wildly divergent policies ($\Gamma, \gamma, \mathcal{S}, \mathcal{P}$, etc) as discussed above.
    \section{Experiments}
\label{secs:experiments}

In these experiments, we consider the behaviour of arbitrarily sized teams of agents ($R \in [2, 6]$) which run as processes on hardware (limited to $4$ cores and $16$GB RAM) that share no physical medium except for a network connection (with latencies ranging from \SI{50} to \SI{500}{\ms}).

\subsection{Case studies}
\label{secs:experiments:case_studies}

The robots employ various strategies for choosing trading partners, including
\begin{inparaenum}[(i)]
	\item every team member (\textit{ALL}), or
	\item a random team member (\textit{BANDIT-EXPLORE}), or
	\item the most trusted team member (\textit{BANDIT-EXPLOIT}), or
	\item the most trusted team member, and occasionally a random team member (\textit{BANDIT-EXPLORE-EXPLOIT}), or
	\item a dedicated team member which acts as a central server (\textit{CENTRAL}).
\end{inparaenum} 

Maps are created from imagery within a hand-labelled catalogue of streets and local attractions in Central Oxford, and we consider several illustrative case studies, shown in Table \ref{tables:experiments:studies}, where each categorised catalogue is further resolved into reasonably sized sections which the agents treat as products for appraisal and purchase.

The results shown are aggregated over $M = 5$ \textit{Monte Carlo} randomised trials, so as to avoid bias in the arbitrary assignment of logs to each robot.

\begin{table}
	\captionsetup{font=scriptsize}
	\centering
	\begin{tabular}{| l | l | c | c |}
        \hline
        \textbf{Catalogue}	& \textbf{Category}	& \textbf{Stock items}	& \textbf{Metres}\\
        \hline
      	\textit{ST-ANNES} 		& college & 12 & 143\\
        \hline
        \textit{BEVINGTON} 		& street & 15 & 180\\
        \hline
        \textit{RHODES} 		& house & 21 & 260\\
        \hline
        \textit{TRINITY} 		& college & 28 & 350\\
        \hline
        \textit{BLACKHALL} 		& street & 17 & 210\\
        \hline
        \textit{OBSERVATORY} 	& street & 26 & 322\\
        \hline
        \textit{ORI} 			& lab & 36 & 450\\
        \hline
        \textit{BROAD} 			& street & 18 & 440\\
        \hline
        \textit{MATERIALS} 		& street & 24 & 300\\
        \hline
	\end{tabular}
	\caption{Exemplar demarcation of the \textit{Oxford RobotCar Dataset} \cite{RobotCarDatasetIJRR}, as a portion of the $55$ logs, or \SI{446}{\km}, of which is labelled and categorised. Each catalogue consists of a range of ``stock items'' which correspond to drivable sections of the world in the vicinity of the street or local attraction. Note that in the case of \textit{BROAD}, an approximately evenly sized loop of a \SI{220}{\m} stretch means that this catalogue's product range is less dense than, say, that of \textit{ORI}.}
	\label{tables:experiments:studies}
\end{table}

\subsection{An appreciable measure of localisation performance}
\label{secs:experiments:robustness}

We measure localisation performance by counting drop-outs, $X \in \mathcal{X}$,  (where a robot is travelling using VO alone), and present cumulative failure distributions $P(X \geq x | f) \propto P(X \geq x)$ which represent the probability of a robot having to travel another $x$ metres blindly after having lost localisation.
We understand our measure of success in comparison to the localisation performance of the mapping framework alone (\textit{EBN}).

\subsection{Sharing expertise to bolster localisation performance}
\label{secs:experiments:accuracy}

Our first case study involves a team of $4$ robots set loose onto $4$ distinct streets (\textit{BEVINGTON}, \textit{BROAD}, and \textit{PARKS}) and along the front of a laboratory, \textit{ORI}.
Clearly, trading navigation data is beneficial, where the failure distributions are consistently improved.

\begin{figure}
    \centering
    
    \subfigure[\textit{BEVINGTON} street.]{
    	\includegraphics[scale=0.2]{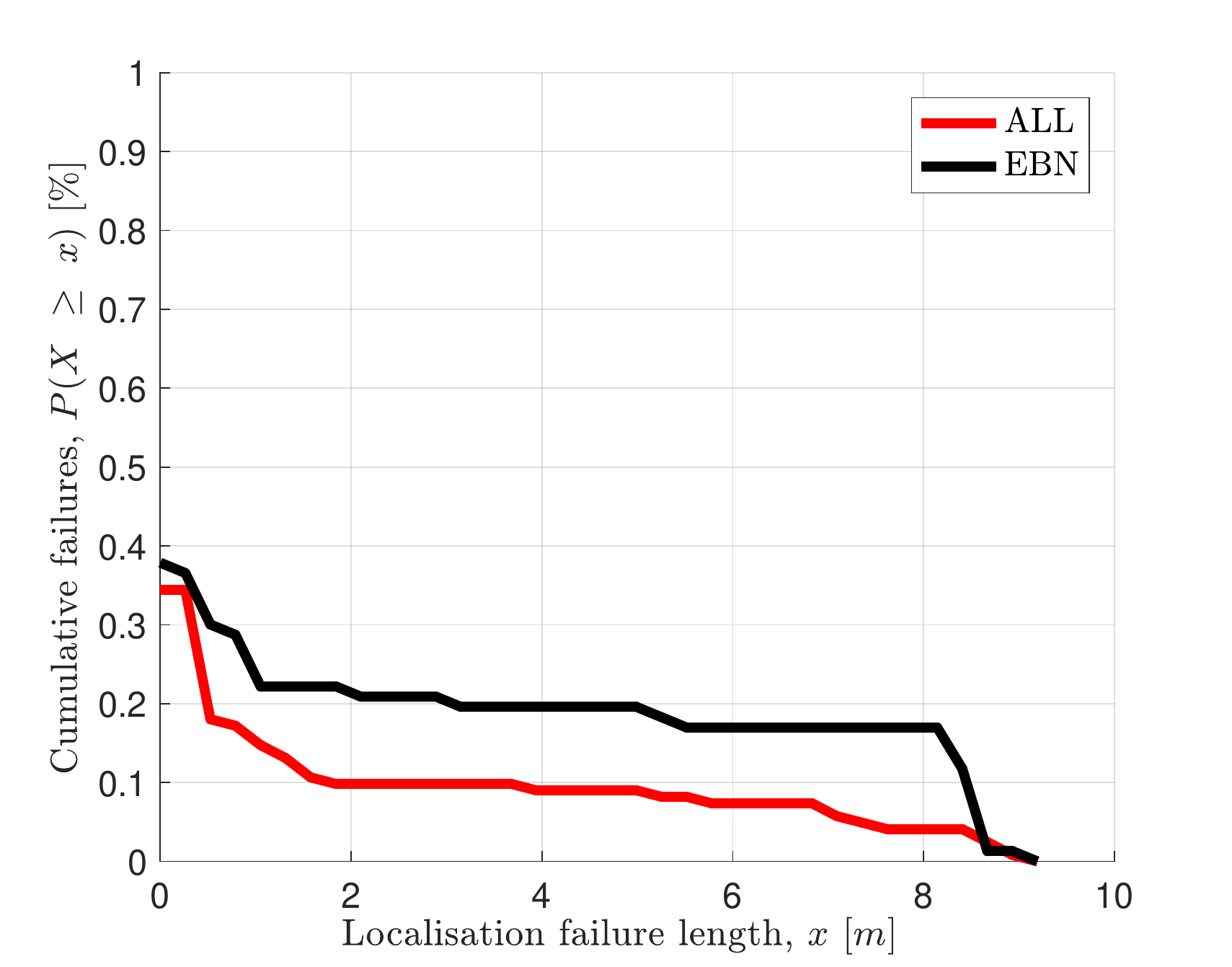}
    	\label{figs:experiments:hope:bevington}
    }
	\quad
    \subfigure[\textit{BROAD} street.]{
    	\includegraphics[scale=0.2]{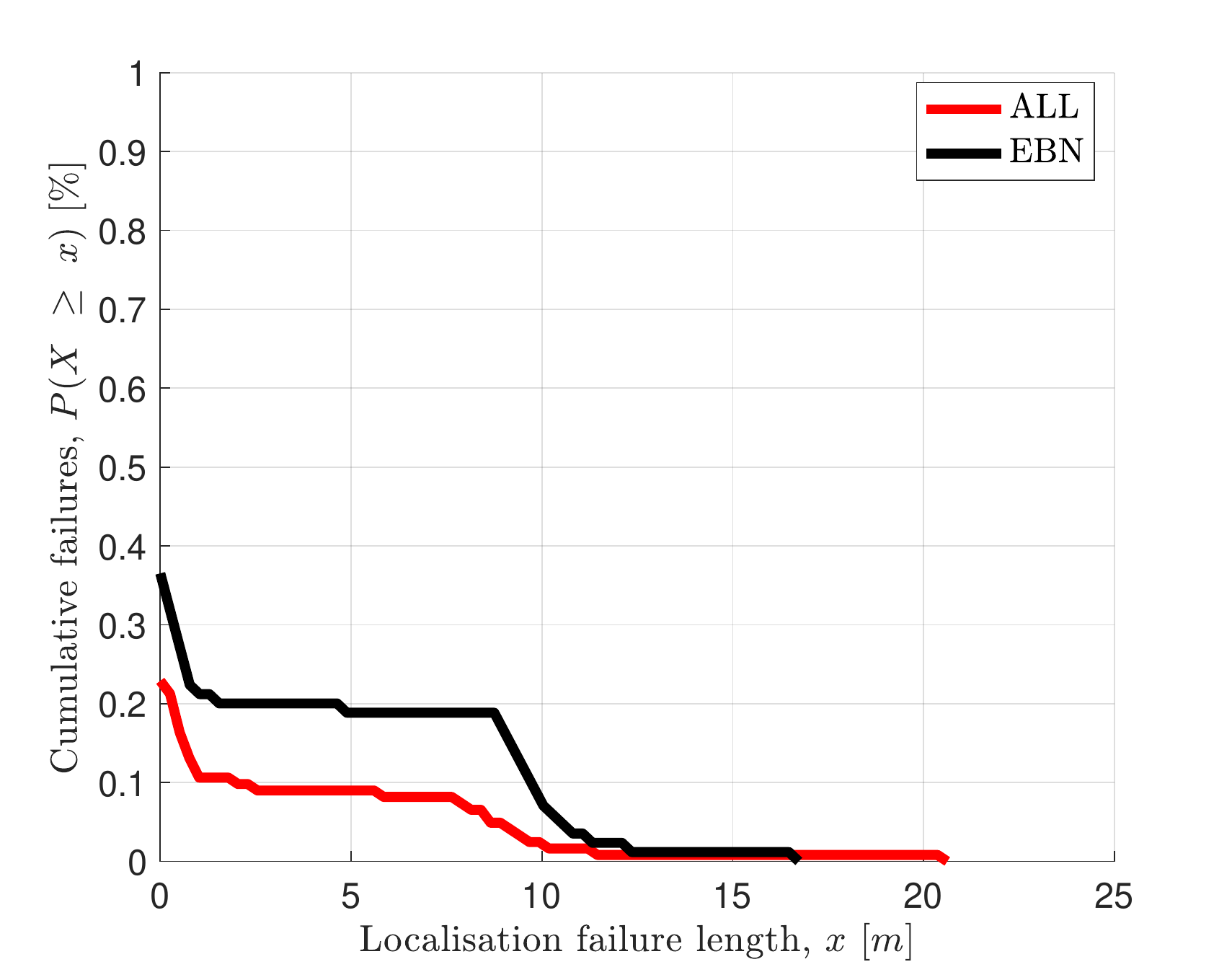}
    	\label{figs:experiments:hope:broad}
    }

    \subfigure[\textit{ORI} laboratory.]{
    	\includegraphics[scale=0.2]{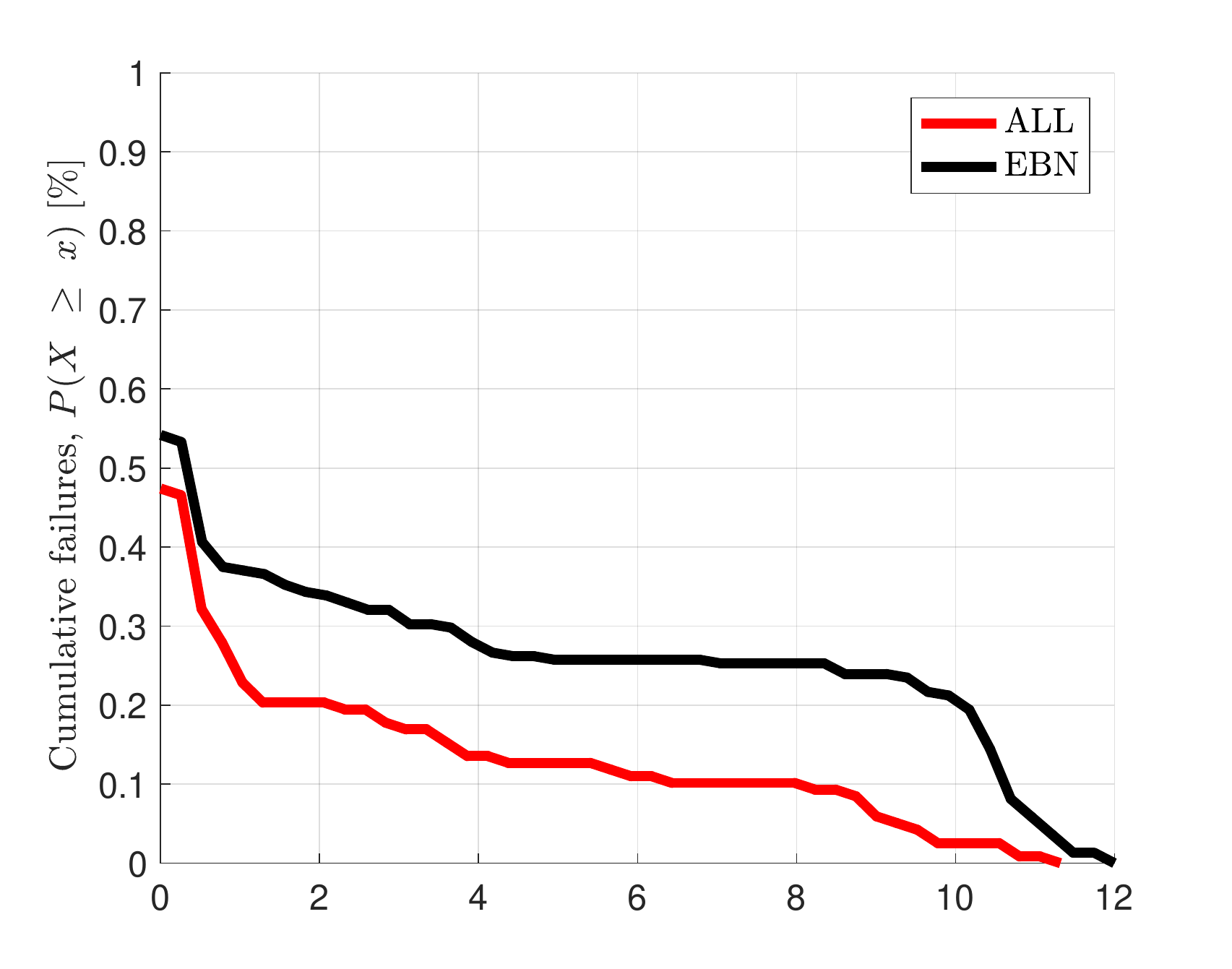}
    	\label{figs:experiments:hope:ori}
    }
	\quad
    \subfigure[\textit{TRINITY} college.]{
    	\includegraphics[scale=0.2]{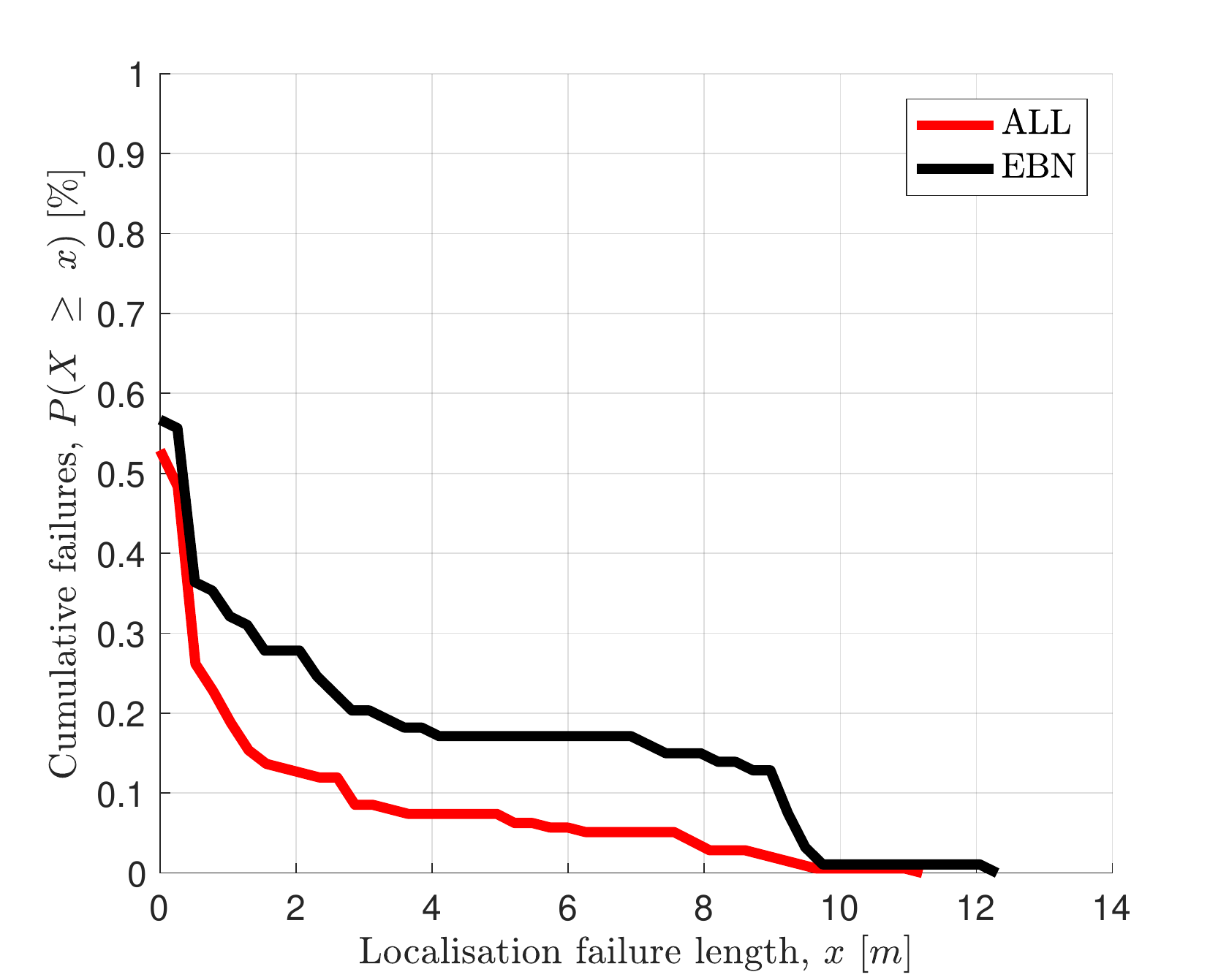}
    	\label{figs:experiments:hope:trinity}
    }

    \caption{Trading navigation data bolsters localisation robustness, shown here by statistics over $4$ distinct sections of the dataset at \cite{RobotCarDatasetIJRR}. For illustrative purposes, we aggregate failures on a per-street basis, however each robot has a diagnostic measure of its performance at a finer resolution, which it uses in instigating trades.}
    \label{figs:experiments:hope}
    \vspace{-.6cm}
\end{figure}

\subsection{Scaling marketplace interactions with the size of the fleet}
\label{secs:experiments:scaling}

\begin{figure}
    \centering

    \subfigure[Average per-robot CPU time spent performing delivering nodes purchased by another agent as a function of the team size, where querying \textit{ALL} agents quickly scales up unreasonably.]{
        \includegraphics[scale=0.2]{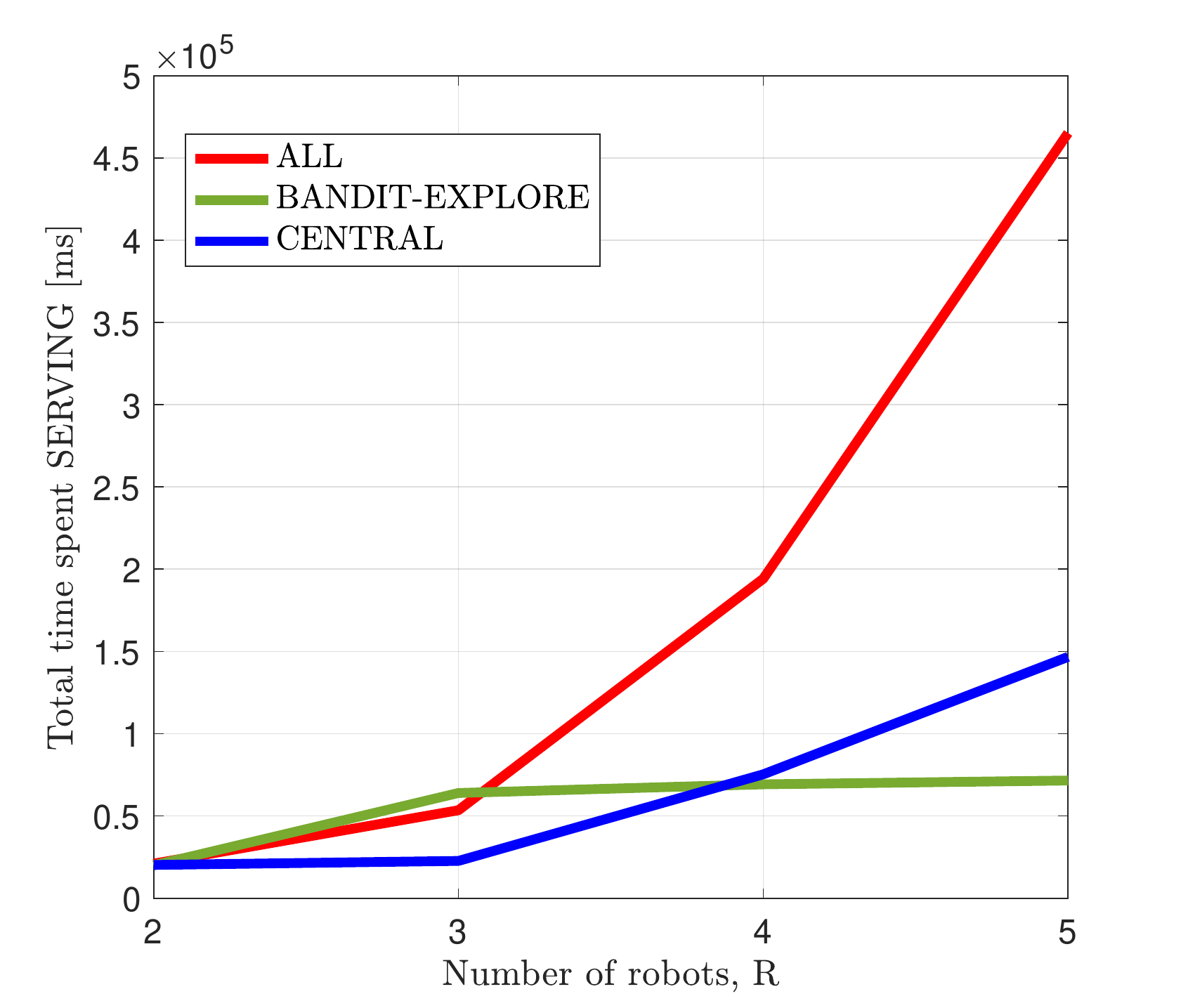}
        \label{figs:experiments:scaling:elapsed_milliseconds}
    }
    \quad
    \subfigure[Average per-robot network load as a function of the team size, measured here as queries sent into the market.]{
        \includegraphics[scale=0.2]{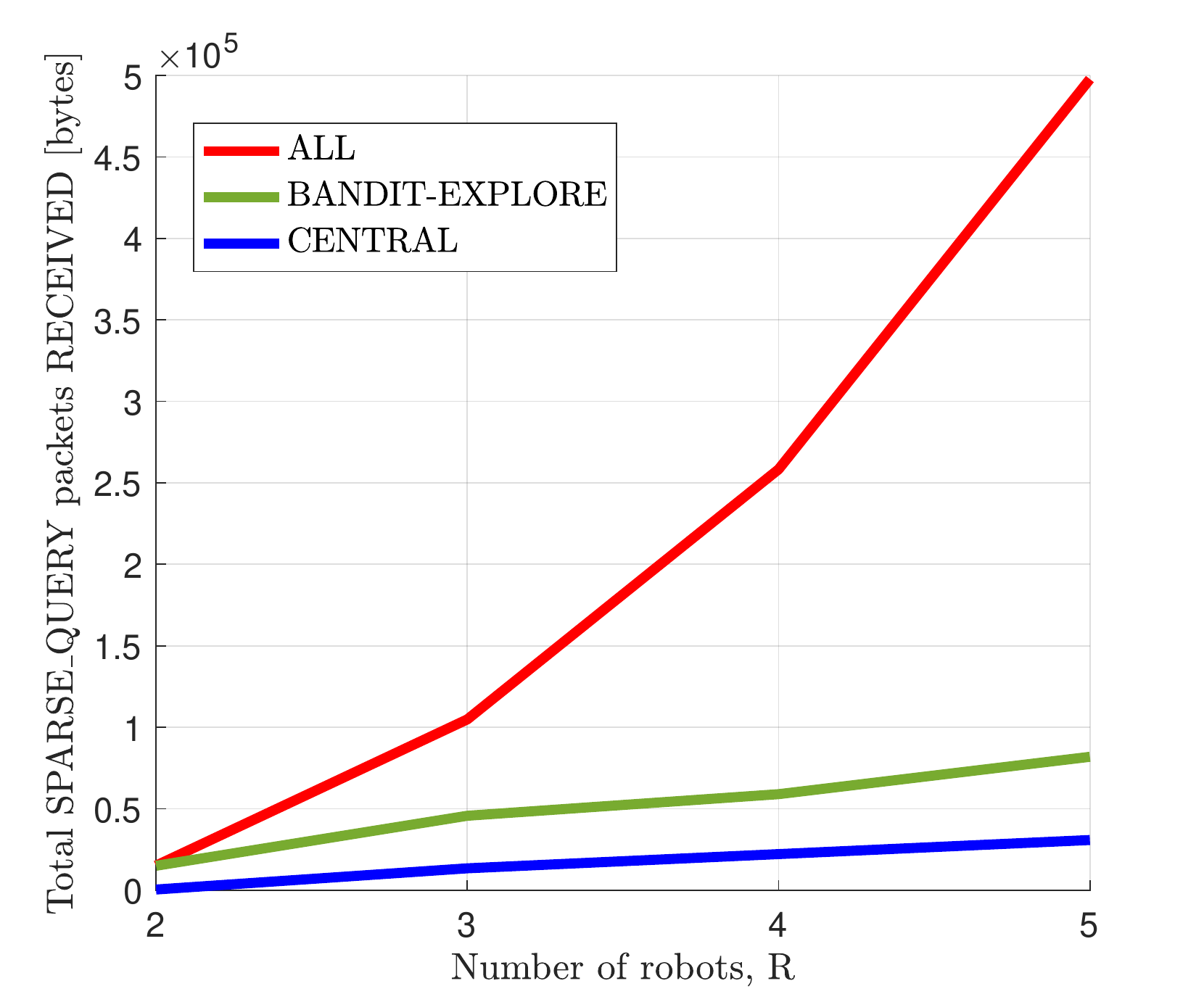}
        \label{figs:experiments:scaling:bytes}
    }
    
    \caption{Investigating performance degradation in multi-robot teams of arbitrary size.}
    \label{figs:experiments:scaling}
\end{figure}

In Figure \ref{figs:experiments:scaling}, we investigate how some trading strategies scale with the size of the team.
Here, teams ranging in membership from $2$ to $5$ robots are regularly mapping and trading the area around \textit{ST-ANNES} College.
In Figures \ref{figs:experiments:scaling:elapsed_milliseconds} and \ref{figs:experiments:scaling:bytes} we see that the average total time spent responding to requests and average total query bytes sent over the network degrades as the size of the team increases, unless a canny selection of trading partner is made. 

\subsection{Mitigating resource limitations}
\label{secs:experiments:resources}

\begin{figure}
	\centering
	
	\subfigure[Boundless accumulation of database content is mitigated by restricting patch updates at each trading opportunity.]{
		\includegraphics[scale=0.2]{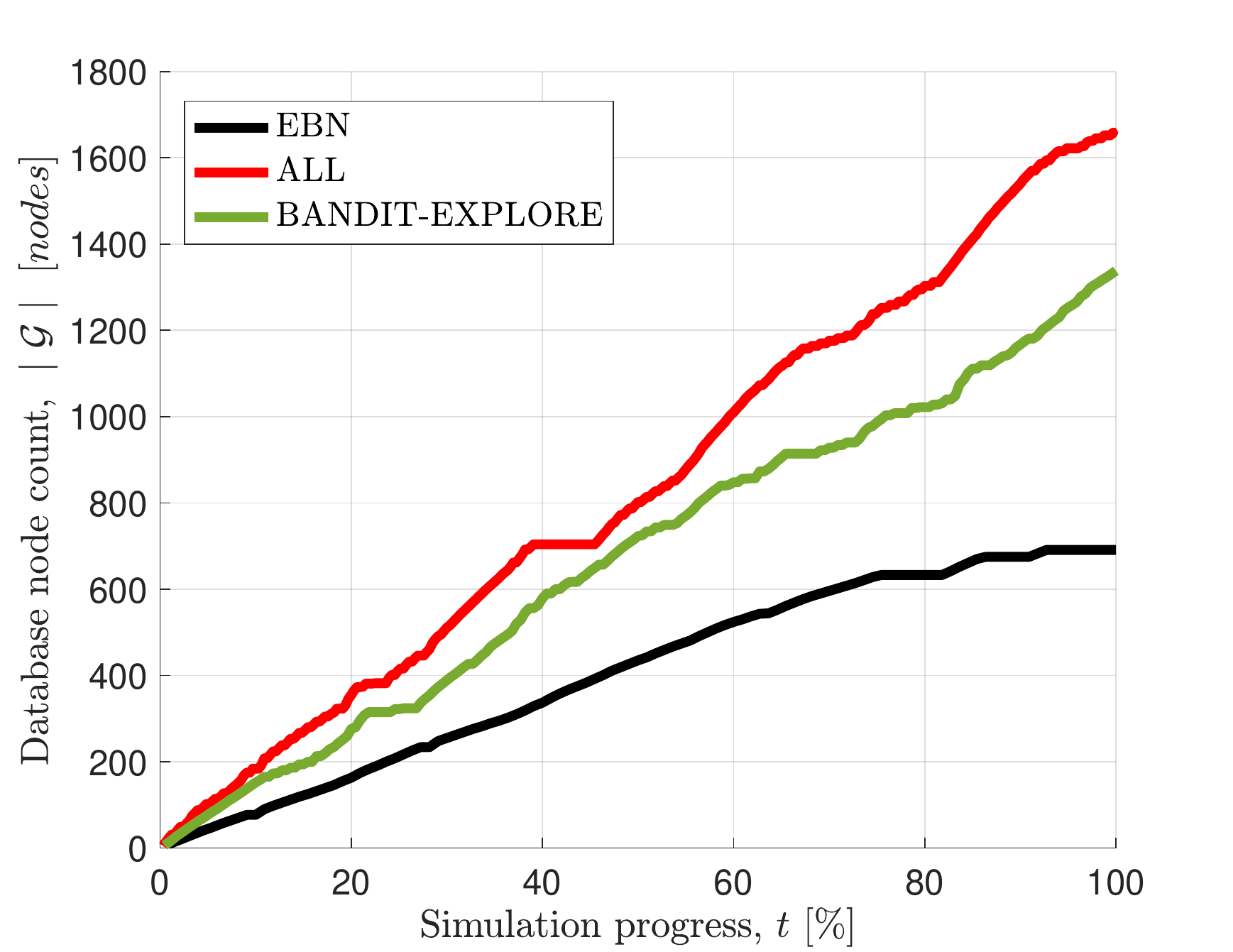}
		\label{figs:experiments:resources:blackhall:log}
	}
	\quad
	\subfigure[Bandwidth restrictions are managed by choosing trading partners with discrimination, requiring fewer node packets to be sent over the network.]{
		\includegraphics[scale=0.2]{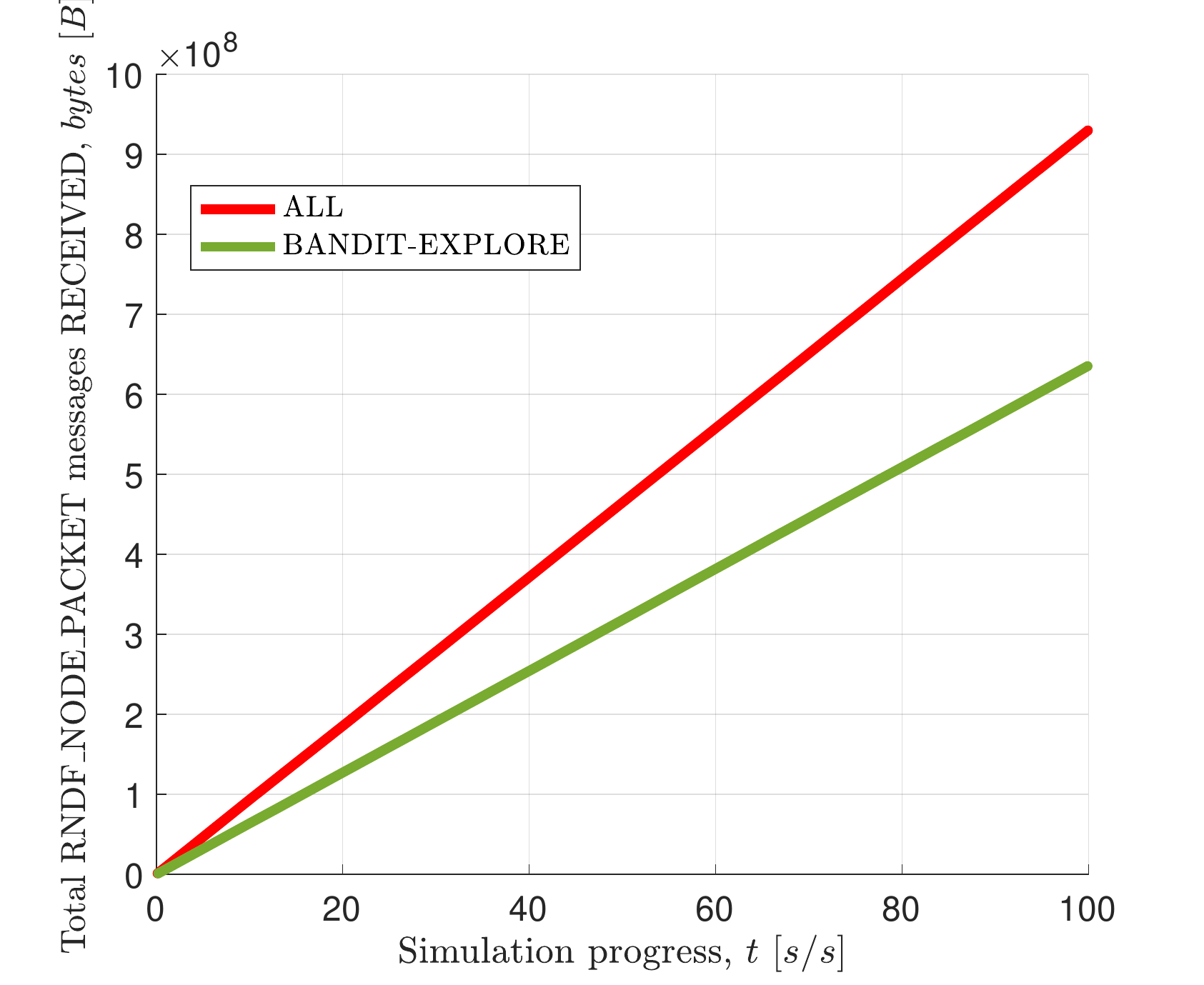}
		\label{figs:experiments:resources:blackhall:bytes_rndf_node_packet_received}
	}
	\quad
	\subfigure[The total CPU time dedicated to downloading data from the market is more lenient under a \textit{BANDIT-EXPLORE} strategy.]{
		\includegraphics[scale=0.2]{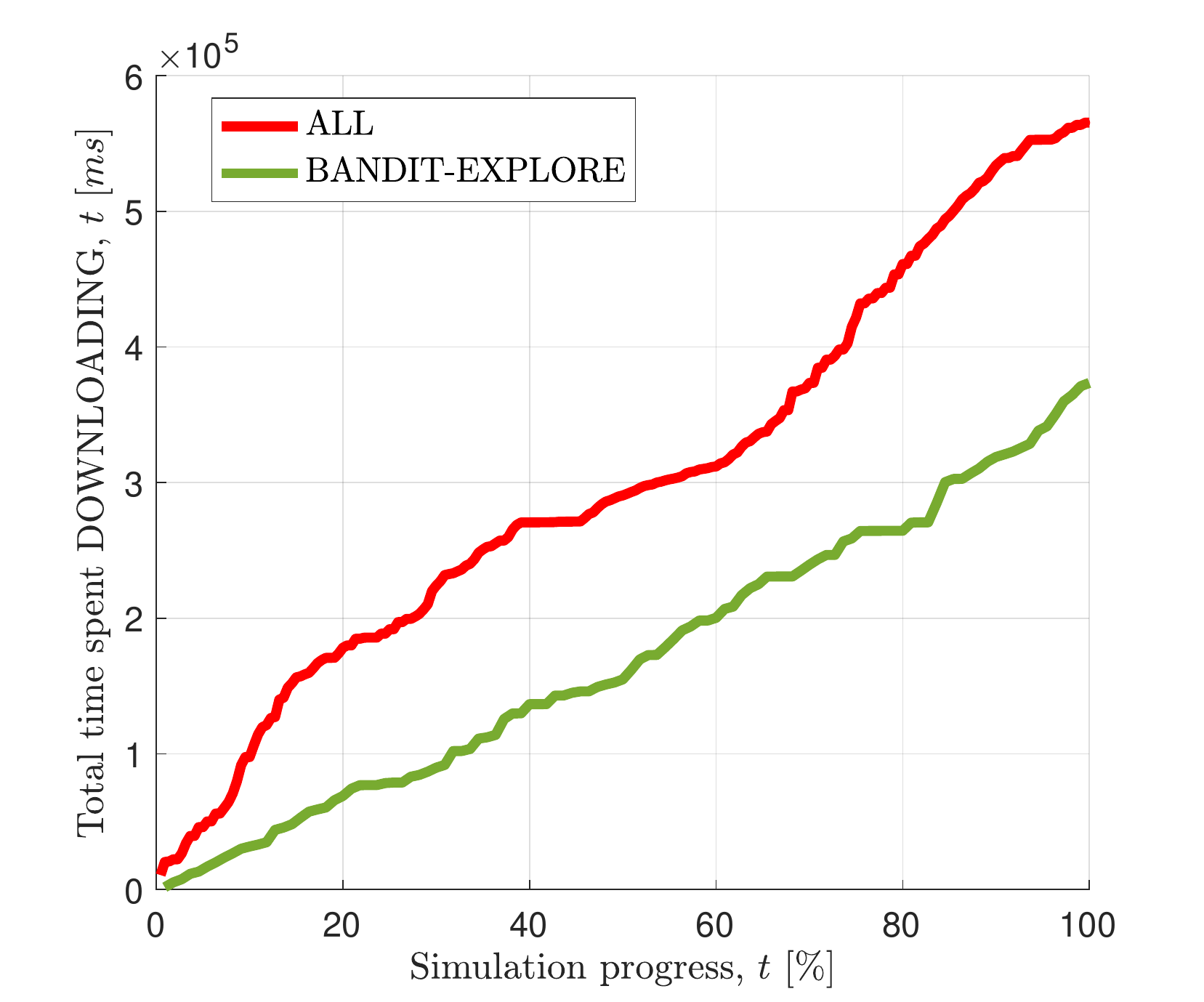}
		\label{figs:experiments:resources:blackhall:elapsed_milliseconds_downloading}
	}
	\quad
	\subfigure[Moderate dropoff in localisation robustness, which is still superior to single-agent EBN.]{
		\includegraphics[scale=0.2]{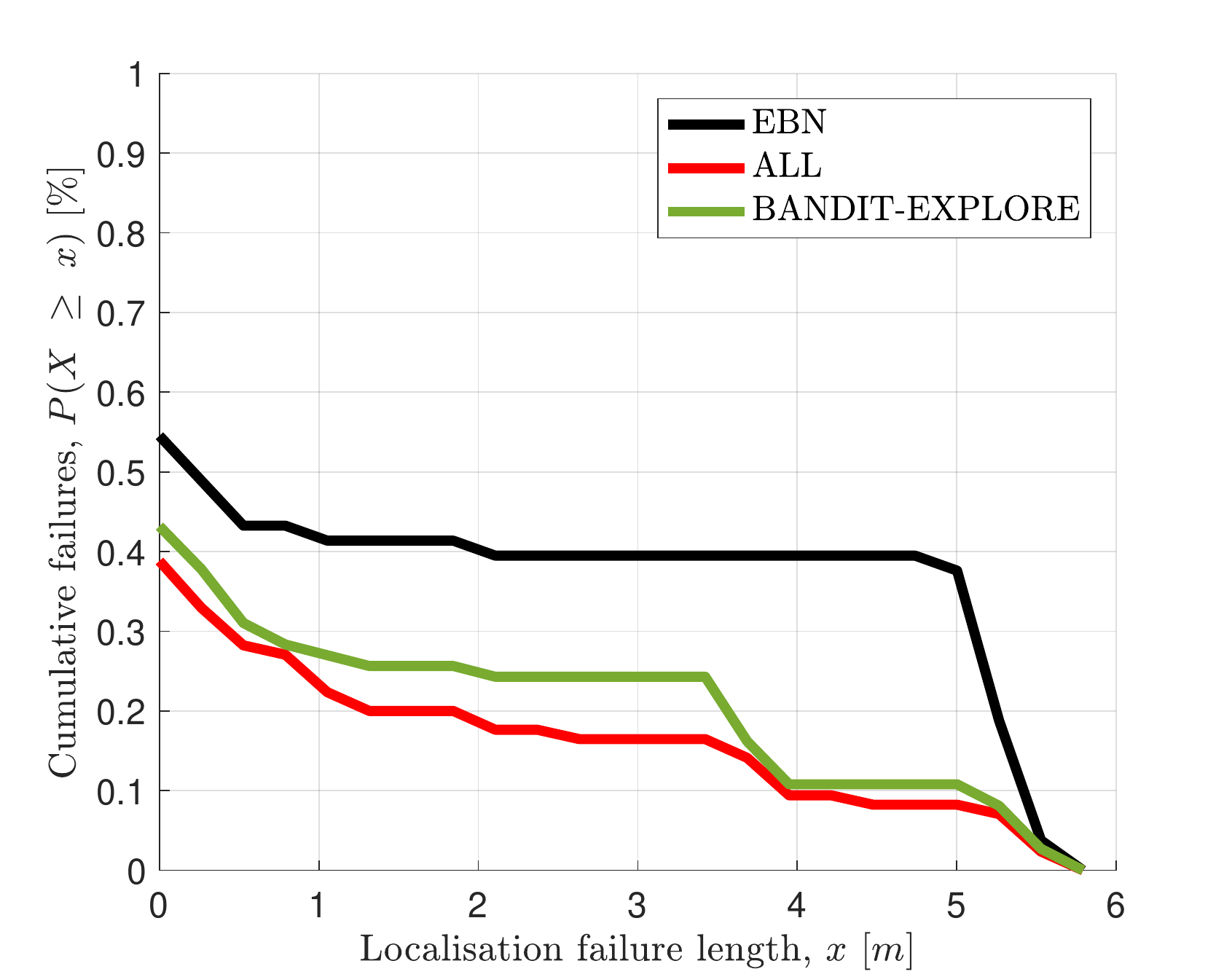}
		\label{figs:experiments:resources:blackhall:cumulative}
	}

	\caption{Resource-limited field robots cannot and should not share navigation data without being equipped with policies for making and accepting queries.}
	\vspace{-.6cm}
	\label{figs:experiments:resources:blackhall}
\end{figure}

The \textit{ALL} trading strategy does not lend itself well to resource-constrained field robots.
We show in Figure \ref{figs:experiments:resources:blackhall} some instrumentation over trials on \textit{BLACKHALL} street. 
Through limiting the range of queries to the rest of the team, we can alleviate map size (Figure \ref{figs:experiments:resources:blackhall:log}), network load (Figure \ref{figs:experiments:resources:blackhall:bytes_rndf_node_packet_received}), and time spent downloading packets (Figure \ref{figs:experiments:resources:blackhall:elapsed_milliseconds_downloading}).
Importantly, we see the expected (but moderate) drop-off in localisation performance (Figure \ref{figs:experiments:resources:blackhall:cumulative}).

\subsection{Exploiting trusted vendors}
\label{secs:experiments:trust}

\begin{figure}
	\centering
	\includegraphics[scale=0.2]{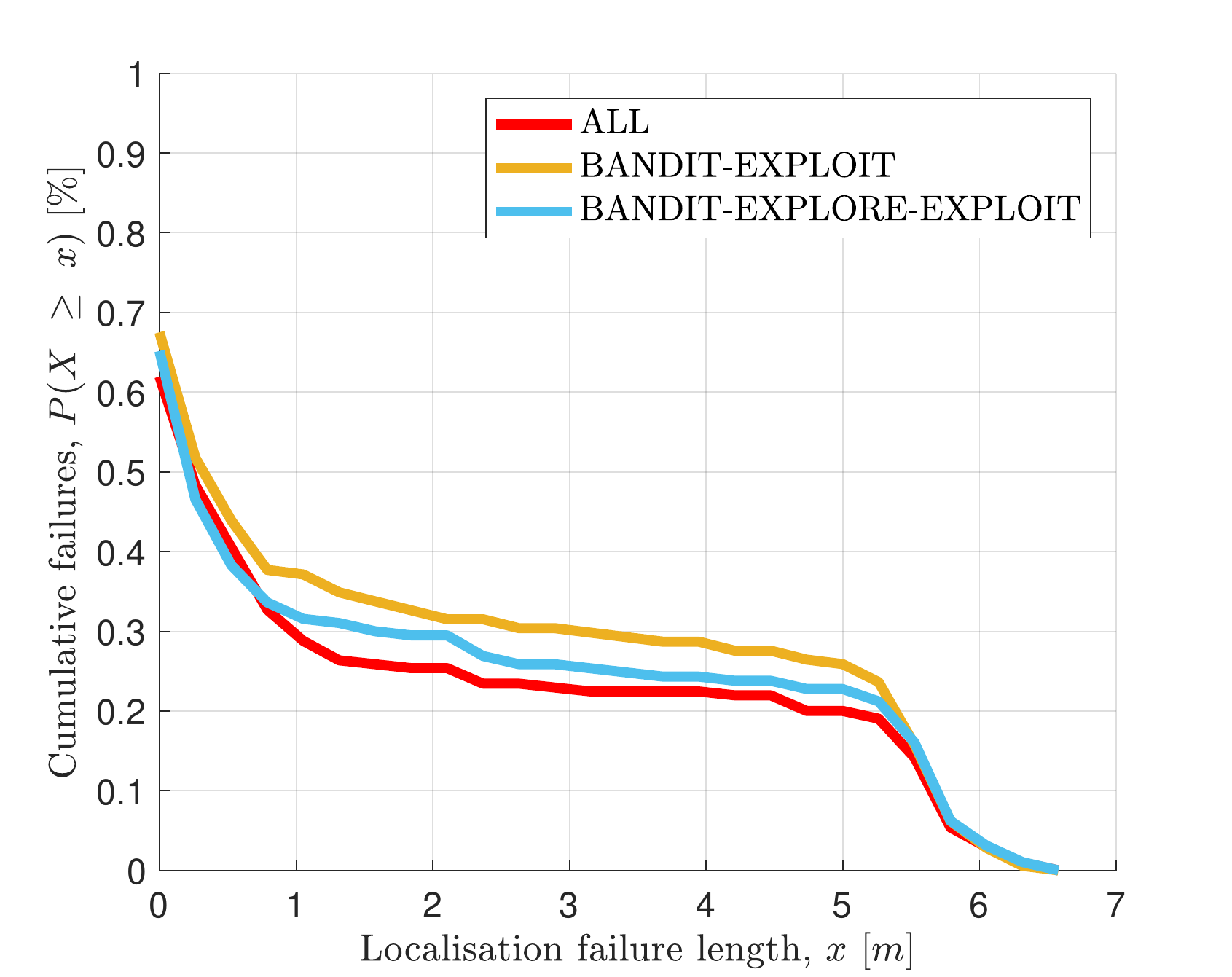}
	\caption{Pure exploitation of the most trusted team member is outperformed by occasionally sampling from another team member, who may have in the interim entered data into the market which is beneficial to the team.}
	\label{figs:experiments:trust:rhodes:cumulative}
	\vspace{-.25cm}
\end{figure}

In Figure \ref{figs:experiments:trust:rhodes:cumulative}, in which a team of $5$ robots operate to the North of \textit{RHODES} house, we find that the most robust strategy for leveraging internally held beliefs is the occasional exploration of a less trustworthy team member, rather than consistently exploiting the favourite seller.
Indeed, this perturbation is required to avoid a local maximum in appreciation for a member of the team, where data variability dictates that some other member will eventually enter data into the market that every other robot will find useful (ephemeral, information-rich appearance change such as snow in Southern England). 

\subsection{Valuing data in the market}
\label{secs:experiments:value}

\begin{figure}
	\centering
	
	\subfigure[The FAB-MAP selection mechanism is more discriminative in appearance than the local \textit{SURF} features used by our VO pipeline.]{
		\includegraphics[scale=0.2]{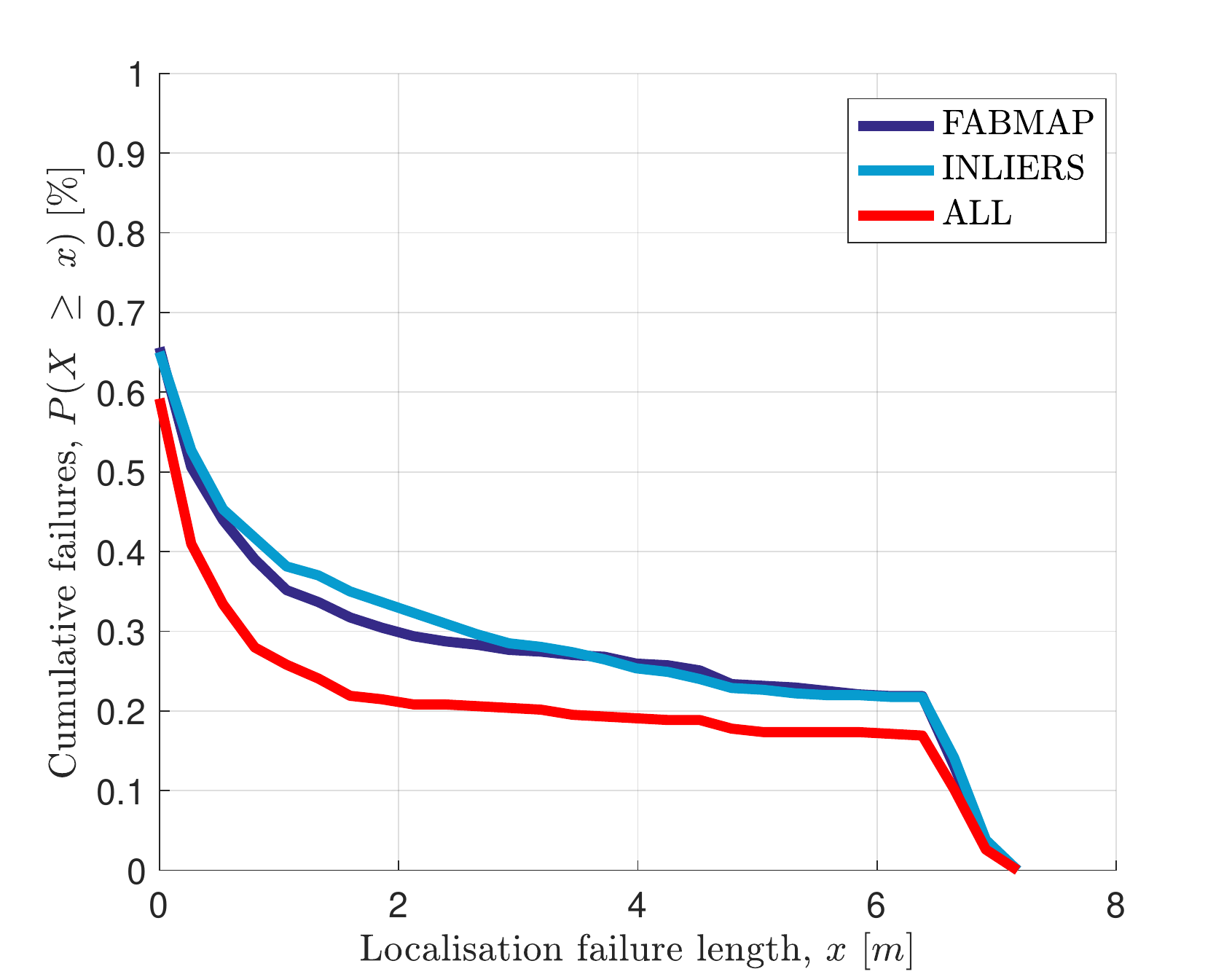}
		\label{figs:experiments:value:observatory:cumulative}
	}
	\quad
	\subfigure[The use of a value metric as a selection mechanism reduces the average sizes for the candidate node set used by the navigation suite for matching foreign content.]{
		\includegraphics[scale=0.2]{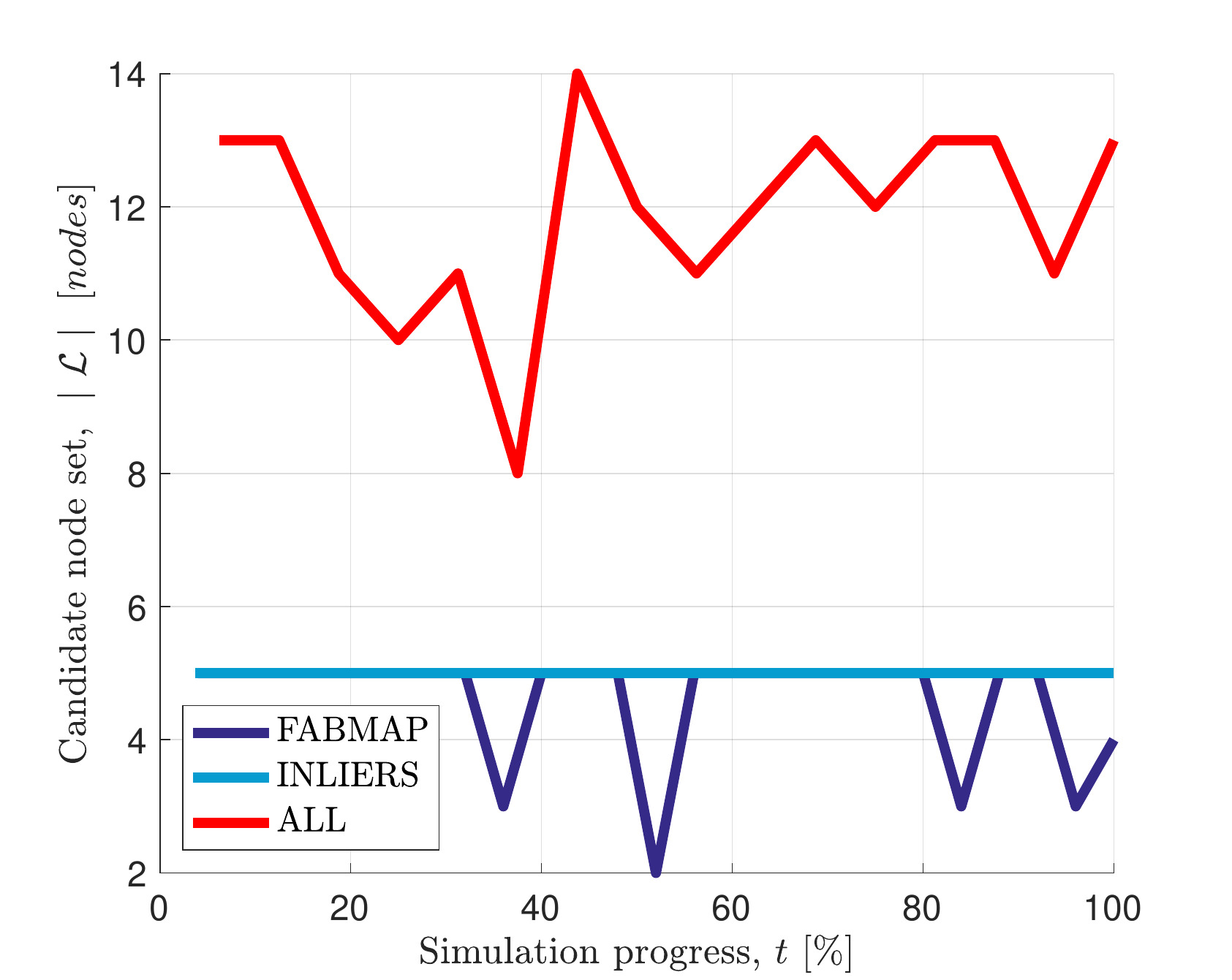}
		\label{figs:experiments:value:observatory:num_dm_patch_elements_lhs}
	}

	\caption{Using a value metric for patches received from agents in the marketplace is beneficial to real-time performance.}
	\label{figs:experiments:value:observatory}
	\vspace{-.6cm}
\end{figure}

In Figure \ref{figs:experiments:value:observatory}, a team of $3$ robots is trading within \textit{OBSERVATORY} street, using various pricing strategies for valuing data in the market, which is further used to dictate
\begin{inparaenum}[(i)]
	\item a limit on packets sent over the network during exchanges,
	\item bounds for the local graph neighbourhood that the localiser matches foreign content within, and
	\item the selection mechanism for closely-matching content. 
\end{inparaenum}	
This feature significantly reduces computational load, where we show in Figures \ref{figs:experiments:value:observatory:num_dm_patch_elements_lhs} the aggregated left-hand-side (LHS) to right-hand-side (RHS) nodes that the underlying localiser attempts to match.
In Figure \ref{figs:experiments:value:observatory:cumulative}, we find that the best metric we are currently equipped with is the normalised match scores using \textit{FAB-MAP} vocabulary queries.

\subsection{Managing a useful shopping list}
\label{secs:experiments:shopping}

\begin{figure}
	\centering
	\includegraphics[scale=0.2]{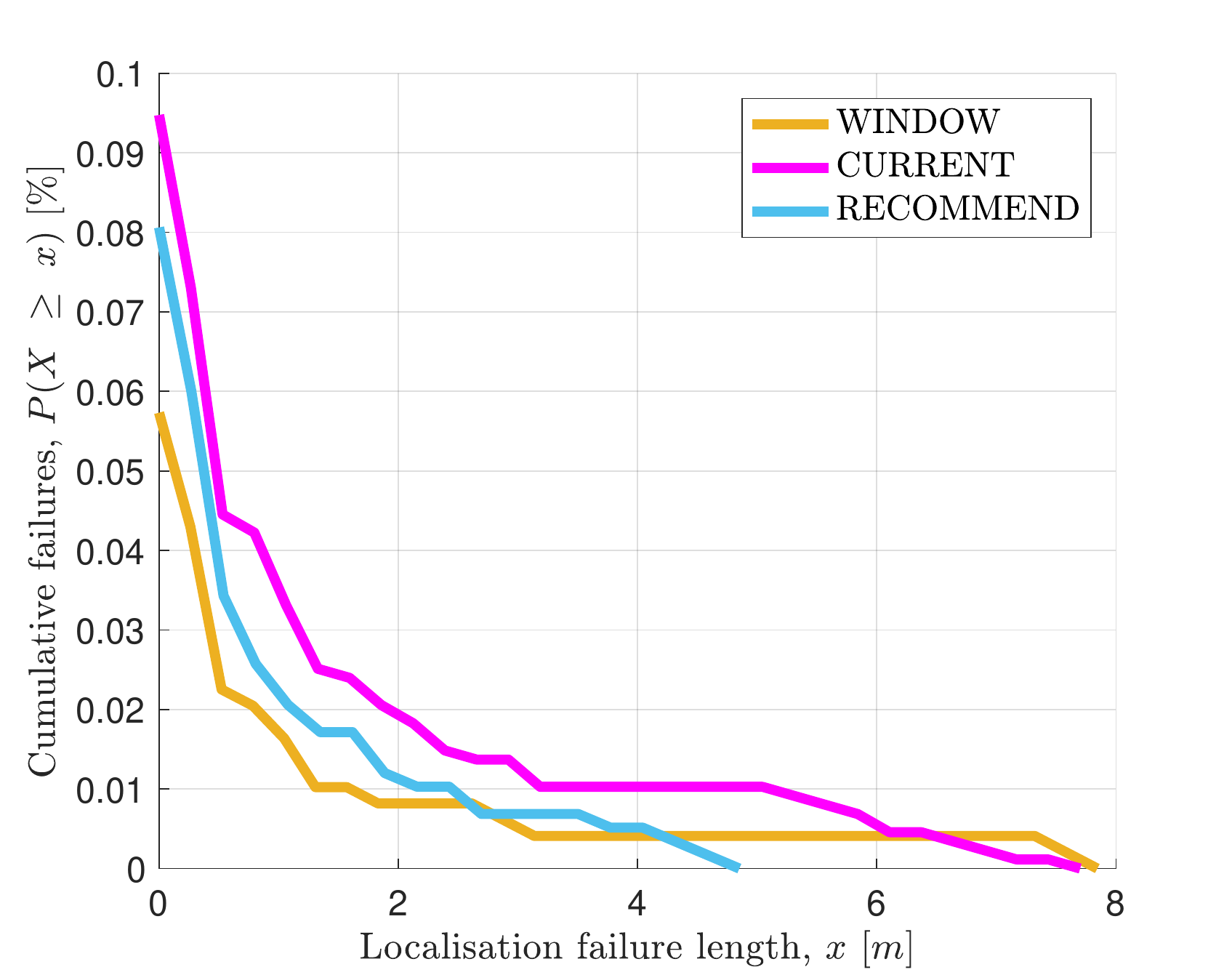}
	\caption{Intelligent shopping strategies are beneficial to localisation performance in general, where both \textit{WINDOW} and \textit{RECOMMEND} strategies outperform the simplest method for querying catalogue categories. \textit{WINDOW} performs best for small failures (an agent queries a padded envelope around its current weakly localised pose), but the \textit{RECOMMEND} strategy has the shortest tail, and as such has consistently prevented the most catastrophic of failures.}
	\label{figs:experiments:shopping:materials:cumulative}
	\vspace{-.6cm}
\end{figure}

Finally, we illustrate the flexibility of our system in allowing agents in the team to be creative in exploring the product range within the data market.
Here, a team of $4$ robots is mapping the length of the street adjacent to the \textit{MATERIALS} building.
In Figure \ref{figs:experiments:shopping:materials:cumulative} we see that the simplest strategy, purchasing the product \textit{CURRENT} placed within or mapping on, is outperformed by including on the shopping list
\begin{inparaenum}[(i)]
	\item products in a \textit{WINDOW} adjacent to \textit{CURRENT} within the product policy $\mathcal{P}$, and
	\item products which other team members \textit{RECOMMEND}.
\end{inparaenum}
    \section{Conclusion}
\label{secs:conclusion}

We have presented our work over the last year towards a totally decentralised versioning system for experience maps in which teams of miscellaneous, resource-starved robots engaging in regular forays into a complex world can leverage each others navigation expertise to deal with challenging modes of appearance change.
We show through a set of illustrative case-studies, evaluated over several representative attractions within \SI{446}{\km} of stereo imagery, the boons our framework offers to localisation performance, as well its more lenient demands on network, disk, and CPU resources, particularly as the size of the team increases.
Our ``data market'' presents itself as a landscape for algorithms, and is accessible to agents with variegated configurations and competing interests.
    \section*{Acknowledgements}
\label{secs:acknowledgements}
Matthew Gadd is supported by the FirstRand Laurie Dippenaar, Oppenheimer Memorial Trust, and Ian Palmer scholarships.
Paul Newman is supported by EPSRC Leadership Fellowship Grant EP/J012017/1 and EPSRC Programme Grant EP/M019918/1.
    
    \bibliographystyle{IEEEtran}
    \bibliography{2018ARXIV_gadd}
\end{document}